\documentclass{article}
\usepackage[accepted]{icml2022}

\usepackage[color,cmtip,all]{xy}

\usepackage{hyperref}       
\usepackage[textsize=tiny]{todonotes}

\usepackage{tikz}
\usepackage{microtype}
\usepackage{graphicx}
\usepackage{subfigure}
\usepackage{booktabs}
\usepackage[utf8]{inputenc} 
\usepackage[T1]{fontenc}    

\usepackage{url}            
\usepackage{booktabs}       
\usepackage{amsfonts}       
\usepackage{nicefrac}       
\usepackage{microtype}      
\usepackage{xcolor}   

\usepackage{amssymb}
\usepackage{amsmath}
\usepackage{amsthm}
\theoremstyle{plain}
\usepackage{bbm}
\usepackage{mathtools}
\usepackage{bm}
\usepackage{csvsimple}
\usepackage{dirtytalk}
\usepackage{mdframed}
\usepackage{ulem}
\usepackage{tikz}
\usetikzlibrary{arrows,backgrounds}
\usepgflibrary{shapes.multipart}

\usepackage{graphicx}
\usepackage{mdframed}
\usepackage{stmaryrd} 
\usepackage{pifont}
\usepackage{cancel}

\newcommand{\avoir}{\texttt{a}}

\newcommand{\Ridge}{\texttt{Ridge}}

\newcommand{\GD}{\texttt{GD}}

\newcommand{\Yperm}{\textbf{Y}_{\texttt{perm}}}

\newcommand{\trainmet}{\texttt{AdaCap}}
\newcommand{\shrinkage}{\texttt{Tikhonov}}
\newcommand{\Adaboost}{\texttt{Adaboost}}
\newcommand{\CV}{\texttt{CV}}
\newcommand{\SELU}{\texttt{SeLU}}
\newcommand{\SeLU}{\texttt{SeLU}}
\newcommand{\GLU}{\texttt{GLU}}
\newcommand{\ResBlock}{\texttt{ResBlock}}
\newcommand{\MLP}{\texttt{MLP}}

\newcommand{\Fast}{\texttt{Fast}}
\newcommand{\Linear}{\texttt{Linear}}
\newcommand{\regularnetfast}{\Fast\MLP}
\newcommand{\regularnetstandard}{\MLP}
\newcommand{\regularnetresblock}{\ResBlock }
\newcommand{\regularnetbatchstandard}{\Batch\MLP}
\newcommand{\regularnetbatchresblock}{\Batch\ResBlock}
\newcommand{\regularnetglu}{\GLU{} \MLP}
\newcommand{\regularnetselu}{\SNN{}}
\newcommand{\regularnetfastselu}{\Fast \SNN{}}

\newcommand{\mlrnetstandard}{\trainmet{}\MLP}
\newcommand{\mlrnetresblock}{\trainmet{}\ResBlock}

\newcommand{\mlrnetbatchresblock
}{\trainmet{}\Batch\ResBlock}
\newcommand{\mlrnetglu}{\trainmet{}\GLU{}\MLP}
\newcommand{\mlrnetselu}{\trainmet{}\SNN{} }
\newcommand{\mlrnetfastselu}{\trainmet{}\Fast\SNN{} }

\newcommand{\binclf}{\texttt{BinClf}}
\newcommand{\multiclf}{\texttt{MultiClf}}
\newcommand{\reg}{\texttt{Reg}}

\newcommand{\lr}{\textit{l.r.}}
\newcommand{\Adam}{\texttt{Adam}}

\newcommand{\dropout}{\texttt{DO}}
\newcommand{\batchnorm}{\texttt{BN}}
\newcommand{\Batch}{\texttt{Batch}}

\newcommand{\isTopOne}{\texttt{Top1}}
   
\newcommand{\WD}{{\text{Weight Decay}}}
\newcommand{\SNR}{{\text{SNR}}}
\newcommand{\TD}{{\text{TD}}}
\newcommand{\Rdeux}{{\texttt{R}^2}}

\newcommand{\SVM}{{\texttt{SVM}}}
\newcommand{\DL}{{\text{DL}}}

\newcommand{\GBDT}{{\texttt{GBDT}}}

\newcommand{\TabNet}{{\texttt{TabNet}}}
\newcommand{\DNN}{{\texttt{DNN}}}
\newcommand{\NODE}{{\texttt{NODE}}}
\newcommand{\Catboost}{{\texttt{CatBoost}}}
\newcommand{\Fastcat}{\texttt{FastCat}}
\newcommand{\XGBoost}{{\texttt{XGBoost}}}
\newcommand{\XGB}{{\texttt{XGB}}}

\newcommand{\lightgbm}{{\texttt{LightGBM}}}

\newcommand{\CNN}{{\texttt{CNN}}}

\newcommand{\SNN}{{\texttt{SNN}}}

\newcommand{\NAN}{{\texttt{NAN}}}
\newcommand{\NuSVM}{{\texttt{NuSVM}}}
\newcommand{\Kernel}{{\texttt{Kernel}}}
\newcommand{\LogReg}{{\texttt{LogReg}}}
\newcommand{\CART}{{\texttt{CART}}}

\newcommand{\AutoML}{{\texttt{AutoML}}}

\newcommand{\AUC}{{\texttt{AUC}}}
\newcommand{\Acc}{{\texttt{Acc.}}}

\newcommand{\RF}{{\texttt{RF}}}
\newcommand{\XRF}{{\texttt{XRF}}}
\newcommand{\Enet}{{\texttt{Elastic-Net}}}

\newcommand{\BN}{{\texttt{BN}}}
\newcommand{\DO}{{\texttt{DO}}}

\newcommand{\bs}{b_s}

\newcommand{\BCE}{{\texttt{BCE}}}
\newcommand{\sigmoid}{{\texttt{Sigmoid}}}
\newcommand{\Id}{{\texttt{Id}}}

\newcommand{\MLR}{{\texttt{MLR}}}

\newcommand{\MARS}{{\texttt{MARS}}}

\newcommand{\GLM}{{\texttt{GLM}}}

\newcommand{\ReLu}{{\texttt{ReLU}}}
\newcommand{\FFNN}{{\texttt{FFNN}}}
\newcommand{\MSE}{{\texttt{MSE}}}
\newcommand{\RMSE}{{\texttt{RMSE}}}

\newcommand{\CE}{\texttt{CE}}
\newcommand{\logsoftmax}{\texttt{logsoftmax}}

\newcommand{\iter}{{\texttt{Iter}}}

\newcommand{\Set}{\textbf{\texttt{set}}}

\newcommand{\fR}{\textbf{{function-T}}}

\newcommand{\bA}{\textbf{A}}
\newcommand{\bH}{\textbf{H}}

\newcommand{\bx}{\textbf{x}}
\newcommand{\bu}{\textbf{u}}

\newcommand{\bY}{\textbf{Y}}

\newcommand{\bW}{\textbf{P}}

\newcommand{\bv}{\textbf{v}}

\newcommand{\bxi}{\boldsymbol{\xi}}

\newcommand{\bla}{\lambda}
\newcommand{\bth}{\boldsymbol{\theta}}

\newcommand{\E}{\mathbb{E}}

\newcommand{\R}{\mathbb{R}}

\newcommand{\cG}{\ensuremath{\mathcal{G}}}

\newcommand{\actout}{\texttt{act}_{\texttt{out}}}

\newcommand{\LAS}{\texttt{Lasso}}

\newcommand{\I}{\ensuremath{\mathbb{I}}}

\makeatletter
\newcommand*\bigcdot{\mathpalette\bigcdot@{.5}}
\newcommand*\bigcdot@[2]{\mathbin{\vcenter{\hbox{\scalebox{#2}{$\m@th#1\bullet$}}}}}
\makeatother

\theoremstyle{plain}
\newtheorem{theorem}{Theorem}[section]

\newtheorem{theo}[theorem]{Theorem}
\newtheorem{lemma}[theorem]{Lemma}

\newtheorem{postulat}[theorem]{Fact}

\usepackage{tikz}
\usetikzlibrary{arrows,backgrounds}
\usepgflibrary{shapes.multipart}

\icmltitlerunning{Adaptive Capacity control}

\begin{document}

\twocolumn[
\icmltitle{\trainmet{}: Adaptive Capacity control for Feed-Forward Neural Networks}



\icmlsetsymbol{equal}{*}

\begin{icmlauthorlist}
\icmlauthor{Karim Lounici}{equal,yyy}
\icmlauthor{Katia Meziani}{equal,comp}
\icmlauthor{Benjamin Riu}{equal,yyy}
\end{icmlauthorlist}

\icmlaffiliation{yyy}{CMAP-Ecole Polytechnique\\
  Route de Saclay\\
  91128  PALAISEAU Cedex\\
  FRANCE}
\icmlaffiliation{comp}{CEREMADE - Universit\'e Paris Dauphine-PSL\\
  Place du Maréchal De Lattre De Tassigny\\
75775 PARIS CEDEX 16}

\icmlcorrespondingauthor{Karim Lounici}{karim.lounici@polytechnique.edu}

\icmlkeywords{Deep Learning, Tabular Data, Random permutations}

\vskip 0.3in
]



\printAffiliationsAndNotice{\icmlEqualContribution} 

\begin{abstract}
The {\it capacity} of a ML model refers to the range of functions this model can approximate. It impacts both the complexity of the patterns a model can learn but also {\it memorization}, the ability of a model to fit arbitrary labels. We propose \textbf{Adaptive Capacity} (\trainmet{}), a training scheme for Feed-Forward Neural Networks (\FFNN). \trainmet{} optimizes the $capacity$ of \FFNN{} so it can capture the high-level abstract representations underlying the problem at hand without memorizing the training dataset. \trainmet{} is the combination of two novel ingredients, the \textbf{Muddling labels for Regularization} (\MLR) loss and the \textbf{Tikhonov operator} training scheme. The \MLR{} loss leverages randomly generated labels to quantify the propensity of a model to memorize. We prove that the \MLR{} loss is an accurate in-sample estimator for out-of-sample generalization performance and that it can be used to perform Hyper-Parameter Optimization provided a Signal-to-Noise Ratio condition is met. The \shrinkage{} \texttt{operator} training scheme modulates the $capacity$ of a \FFNN{} in an adaptive, differentiable and data-dependent manner. We assess the effectiveness of \trainmet{} in a setting where \DNN{} are typically prone to memorization, small tabular datasets, and benchmark its performance against popular machine learning methods.
\end{abstract}

\section{Introduction}

Generalization is a central problem in Deep Learning (\DL). It is strongly connected to the notion of {\it capacity} of a model, that is the range of functions a model can approximate. It impacts both the complexity of the patterns a model can learn but also {\it memorization}, the ability of a model to fit arbitrary labels \cite{goodfellow2016deep}. Because of their high capacity, overparametrized Deep Neural Networks (\DNN{}) can memorize the entire train set to the detriment of generalization. Common techniques like Dropout (\DO) \cite{Hinton2012,srivastava14adrop}, Early Stopping \cite{pmlr-v108-li20j}, Data Augmentation \cite{shorten2019survey} or \WD{} \cite{Hanson1988Comparing,krogh1992simple,bos1996using} used during training can reduce the capacity of a \DNN{} and sometimes delay memorization but cannot prevent it \cite{arpit2017closer}.

We propose \trainmet{}, a new training technique for Feed-Forward Neural Networks (\FFNN) that optimizes the $capacity$ of \FFNN{} during training so that it can capture the high-level abstract representations underlying the problem at hand and mitigate memorization of the train set.
\trainmet{} relies on two novel ingredients, the \textbf{\shrinkage{} operator} and the \textbf{Muddling labels Regularization} (\MLR) loss.

\begin{figure}[http!]
\centering
\includegraphics[scale=0.27]{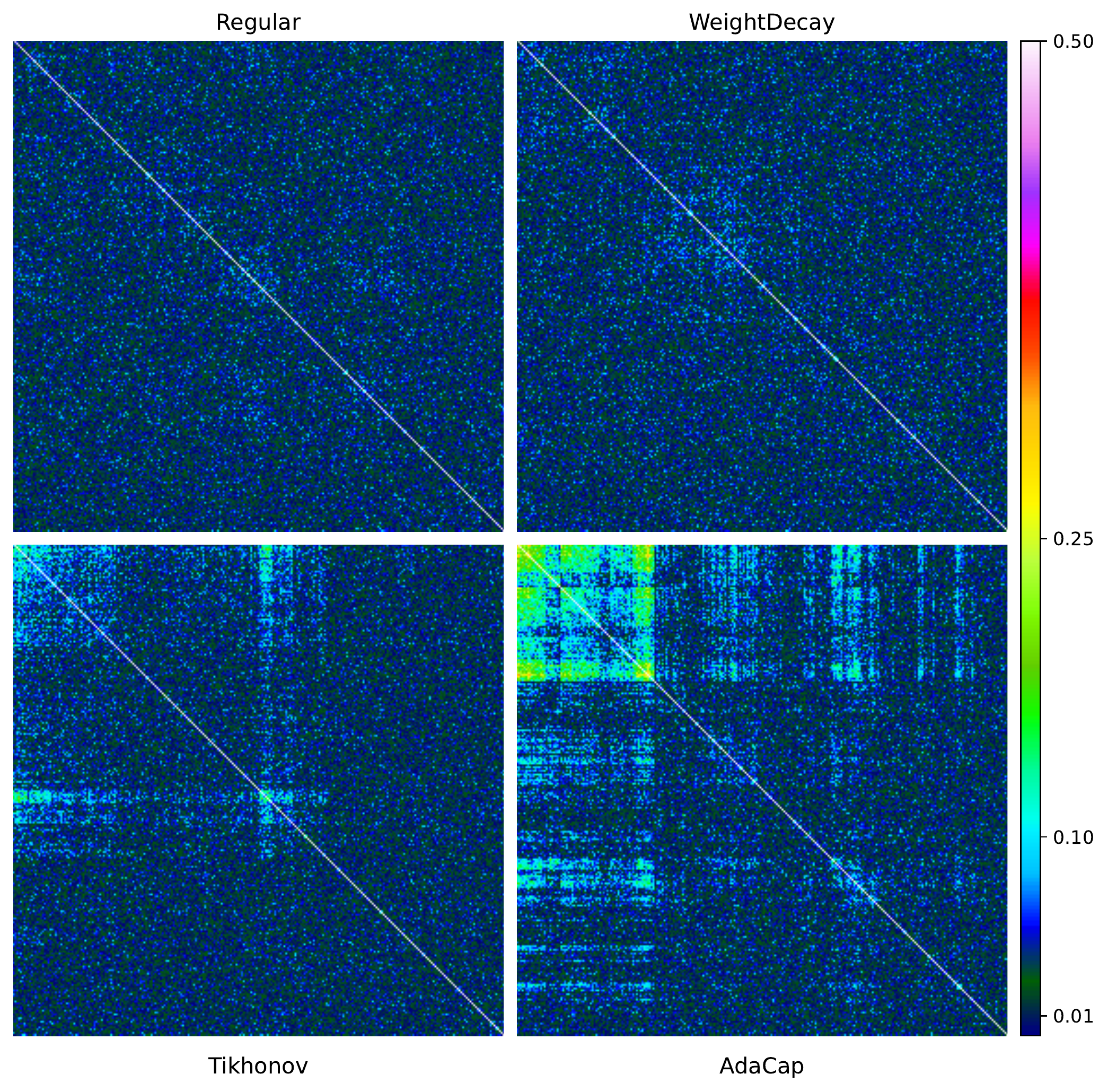}
\caption{We trained a \MLP{} on the \texttt{Boston} dataset with either usual training (with/without \WD{}) or \trainmet{}, using identical architectures and parameters in both cases. We plot the clustered correlation matrices of last hidden layer weights (in absolute value). \textbf{Upper left:} usual loss and no regularization. \textbf{Upper right:} usual loss and \WD{} only. \textbf{Bottom left:} usual loss and \shrinkage{} scheme alone (no \WD). \textbf{Bottom right:} \trainmet{} alone. Contrarily to \WD, \trainmet{} produced a model with highly structured hidden layer weights  even with a plain \MLP{} architecture, indicating its learning behavior is very different from the standard one. } 
\label{fig:1Xcorrelmat}
\end{figure}

The \textbf{\shrinkage{}  
operator} provides a differentiable data-dependent quantification of the capacity of a \FFNN{} through the application of this operator on the output of the last hidden layer. The \shrinkage{} operator modulates the capacity of the \FFNN{} via the additional \shrinkage{} parameter that can be trained concomitantly with the hidden layers weights by Gradient Descent (\GD). This operator works in a fundamentally different way from other existing training techniques like \WD{} (See Section \ref{sec:adapcap} and Fig. \ref{fig:1Xcorrelmat}). 



The problem is then the tuning of the \shrinkage{} parameter that modulates capacity as it directly impacts the generalization performance of the trained \FFNN. This motivated the introduction of the \MLR{} loss which performs capacity tuning without using a hold-out validation set. The \MLR{} loss is based on a novel way to exploit random labels.

Random labels have been used in \cite{zhang2016understanding,arpit2017closer} as a diagnostic tool to understand how overparametrized \DNN{} can generalize surprisingly well despite their capacity to memorise the train set. This benign overfitting phenonemon is attributed in part to the implicit regularization effect of the optimizer schemes used during training \cite{pmlr-v80-gunasekar18a,smith2021on}.
%
Understanding that the training of \DNN{} is extremely susceptible to corrupted labels, numerous methods have been proposed to identify the noisy labels or to reduce their impact on generalization. These approaches include loss correction techniques \cite{patrini2017making}, reweighing samples \cite{jiang2017mentornet}, training two networks in parallel \cite{han2018co}. See \cite{ChenLCZ19,Harutyunyan2020Improving} for an extended survey. 

We propose a different approach. We do not attempt to address the noise and corruptions already present in the original labels. Instead, we \textbf{purposely generate purely corrupted labels during training} as a tool to reduce the propensity of the \DNN{} to memorize label noise during gradient descent. The underlying intuition is that we no longer see generalization as the ability of a model to perform well on unseen data, but rather as the ability to avoid finding pattern where none exists.  
Concretely, we propose the \textbf{Muddling labels Regularization} loss which uses randomly permuted labels to quantify the overfitting ability of a model on a given dataset.  In Section \ref{sec:MLRloss}, we provide theoretical evidences in a regression setting that \textbf{the \MLR{} loss is an accurate estimator of the generalization error} (Fig. \ref{fig:3pred}) which can be used to perform Hyper-Parameter Optimization (HPO) without using a hold-out $validation$-set if a Signal-to-Noise Ratio (\SNR) condition is satisfied.

\begin{figure}[htp]
\centering
\includegraphics[scale=0.35]{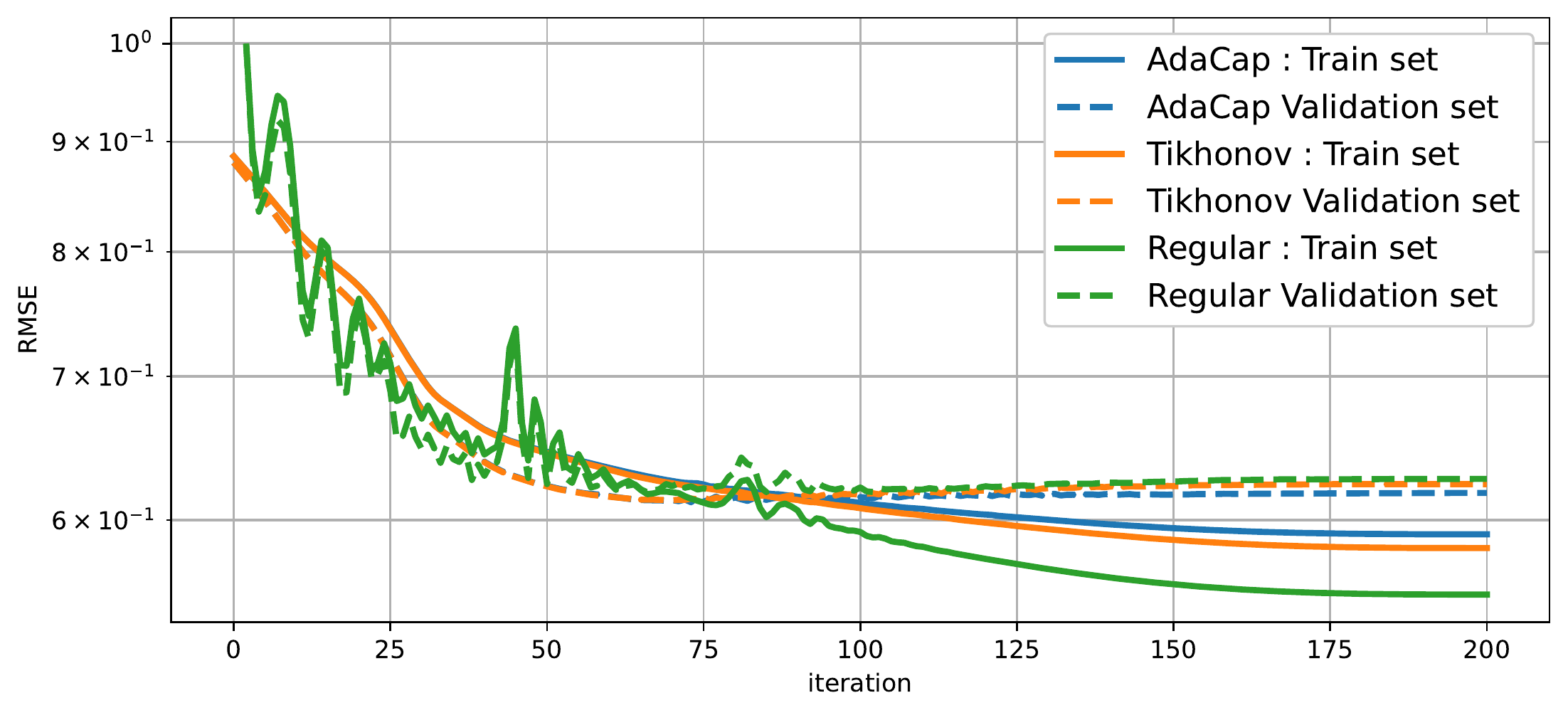}
\caption{\RMSE{} across iterations for train and valid sets when training \trainmet{}, \shrinkage{} and regular \DNN{} on \texttt{Abalone} \TD. We can see the improvement in terms of generalization and memorization when adding \shrinkage{} and then \MLR{}: The train \RMSE{} converges to a higher plateau but validation \RMSE{} keeps improving further. \shrinkage{} smooths the learning dynamic, removes the oscillations. Adding the \MLR{} loss makes no change to the learning dynamic on the first iterations but impact the last iterations and delays memorization even more.}
\label{fig:NewLearningDynamic}
\end{figure}
\begin{figure}[htp]
\centering
\includegraphics[scale=0.43]{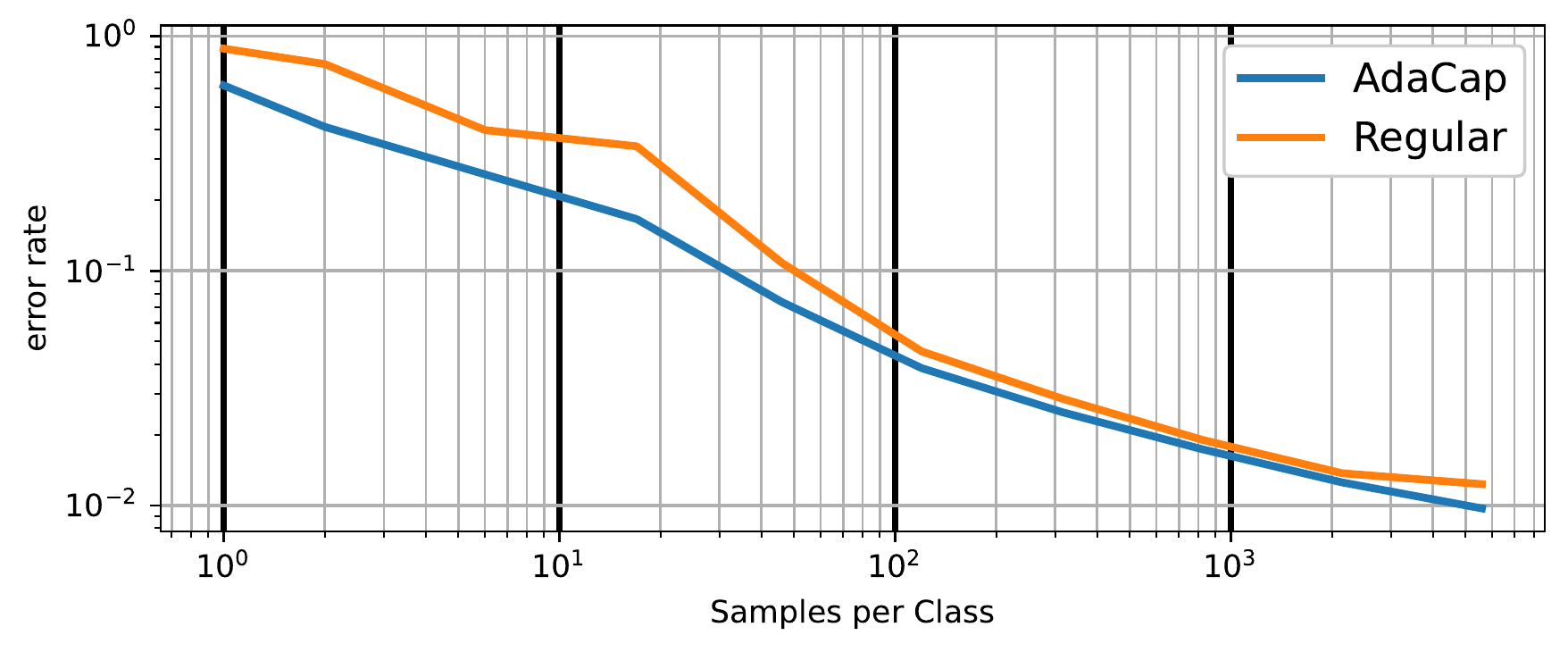}
\caption{Few-shot learning experiment on MNIST \cite{deng2012mnist}. When training a simple ConvNet, the generalization performance of the obtained models $w.r.t.$ the number of samples per class is uniformly better over the whole range of samples per class when using \trainmet{}, especially in the low sample per class regime.
}
\label{fig:fewshot}
\end{figure}

This property motivates using the \MLR{} loss rather than the usual losses during training to perform $adaptive$ control of the $capacity$ of \DNN{}. This can improve generalization (Fig. \ref{fig:NewLearningDynamic}) not only in the presence of label corruption but also in other settings prone to overfitting - $e.g.$ Tabular Data \cite{borisov2021deep,gorishniy2021revisiting,SHWARTZZIV202284}, Few-Shot Learning (Fig. \ref{fig:fewshot}), a task introduced in \cite{fink2005object,fei2006one}. See \cite{wang2020generalizing} for a recent survey.

Our novel training method \trainmet{}  works as follows. Before training: a) generate a new set of completely uninformative labels by {\it muddling} original labels through random permutations; then, at each \GD{} iteration: b) apply the \shrinkage{} operator to the output of the last hidden layer; c) quantify the ability of the  \DNN{}'s output layer to fit true labels rather than permuted labels via the new (\MLR) loss; d) back-propagate the \MLR{} objective through the network.

   
\trainmet{} is a gradient-based, global, data-dependent method which trains the weights and adjusts the capacity of the \FFNN{} simultaneously during the training phase without using a hold-out $validation$ set. \trainmet{} is designed to work on most \FFNN{} architectures and is compatible with the usual training techniques like Gradient Optimizers \cite{kingma2014adam}, Learning Rate Schedulers \cite{smith2019super}, Dropout \cite{srivastava14adrop}, Batch-Norm \cite{ioffe2015batch}, \WD{} \cite{krogh1992simple,bos1996using}, etc.

\DNN{} have not demonstrated yet the same level of success on Tabular Data (\TD) as on images \cite{Krizhevsky2012_cnn}, audio \cite{Hinton2012} and text \cite{bert2019}, which makes it an interesting frontier for \DNN{} architectures. Due to the popularity of tree-based ensemble methods (\Catboost{} \cite{Prokhorenkova2018}, \XGBoost{} \cite{guestrin2016}, \RF{} \cite{barandiaran1998random,breiman2001}), there has been a strong emphasis on the preprocessing of categorical features which was an historical limitation of \DL{}. Notable contributions include \NODE{} \cite{Popov2020Neural} and \TabNet{} \cite{arik2020tabnet}. \NODE{} (Neural Oblivious Decision Ensembles) are tree-like architectures which can be trained end-to-end via backpropagation. \TabNet{} leverages Attention Mechanisms \cite{Bahdanau2014-Attention} to pretrain \DNN{} with feature encoding. The comparison between simple \DNN{}, \NODE, \TabNet, \RF{} and \GBDT{} on \TD{} was made concomitantly by Kadra et~al. \yrcite{kadra2021welltuned}, Gorishniy et~al. \yrcite{gorishniy2021revisiting} and  Shwartz-Ziv \& Armon \yrcite{SHWARTZZIV202284}. Their benchmarks are more oriented towards an \AutoML{} approach than ours, as they all use heavy HPO, and report training times in minutes/hours, even for some small and medium size datasets. See Appendix \ref{app:HPO-bib} for a more detailed discussion. As claimed by \cite{SHWARTZZIV202284}, their results (like ours) indicate that \DNN{} are not (yet?) the alpha and the omega of \TD. \cite{kadra2021welltuned} also introduces an HPO strategy called the regularization cocktail. Regarding the new techniques for \DNN{} on \TD{}, we mention here a few relevant to our work which we included in our benchmark. See \cite{borisov2021deep} and the references therein for a more exhaustive list. \cite{Klambauer2017} introduced Self-Normalizing Networks (\SNN) to train deeper \FFNN{} models, leveraging the \SELU{} activation function. Gorishniy et~al.\yrcite{gorishniy2021revisiting} proposed new architecture schemes: \ResBlock, Gated Linear Units \GLU, and FeatureTokenizer-Transformers, which are adaptation for \TD{} of ResNet \cite{He2015Deep}, Gated convolutional networks \cite{dauphin2017language} and Transformers \cite{vaswani2017attention}.

We illustrate the potential of \trainmet{} on a benchmark of 44 tabular datasets from diverse domains of application, including 26 regression tasks, 18 classification tasks, against a large set of popular methods \GBDT{} \cite{Breiman1997_arcing,Friedman2001,Friedman2002,guestrin2016,Ke2017,Prokhorenkova2018}; Decision Trees and \RF{} \cite{Breiman1984,barandiaran1998random,breiman2001,gey2005model,Klusowski2020sparse}, Kernels \cite{Chihchung2011}, \MLP{} \cite{Hinton89connectionistlearning}, \GLM{} \cite{cox1958,Hoerl1970,tibshirani1996,Zou05}, \MARS{} \cite{Friedman1991}).


For \DL{} architectures, we combined and compared \trainmet{} with \MLP, \GLU, \ResBlock{}, \SNN{} and \CNN. We left out recent methods designed to tackle categorical features (\TabNet, \NODE, \texttt{FT-Transformers}) as it is not the focus of this benchmark and of our proposed method. Our experimental study reveals that using \trainmet{} to train \FFNN{} leads to an improvement of the generalization performance on regression tabular datasets especially those with high Signal-to-Noise Ratio (\SNR), the datasets where it is possible but not trivial to obtain a very small test \RMSE. \trainmet{} works best in combination with other schemes and architectures like \SNN{}, \GLU{} or \ResBlock{}. Introducing \trainmet{} to the list of available \DNN{} schemes allows neural networks to gain ground against the \GBDT{} family. 

\section{The \MLR{} loss} 
\label{sec:MLRloss}

\begin{figure}[htp]
\centering
\includegraphics[scale=0.435]{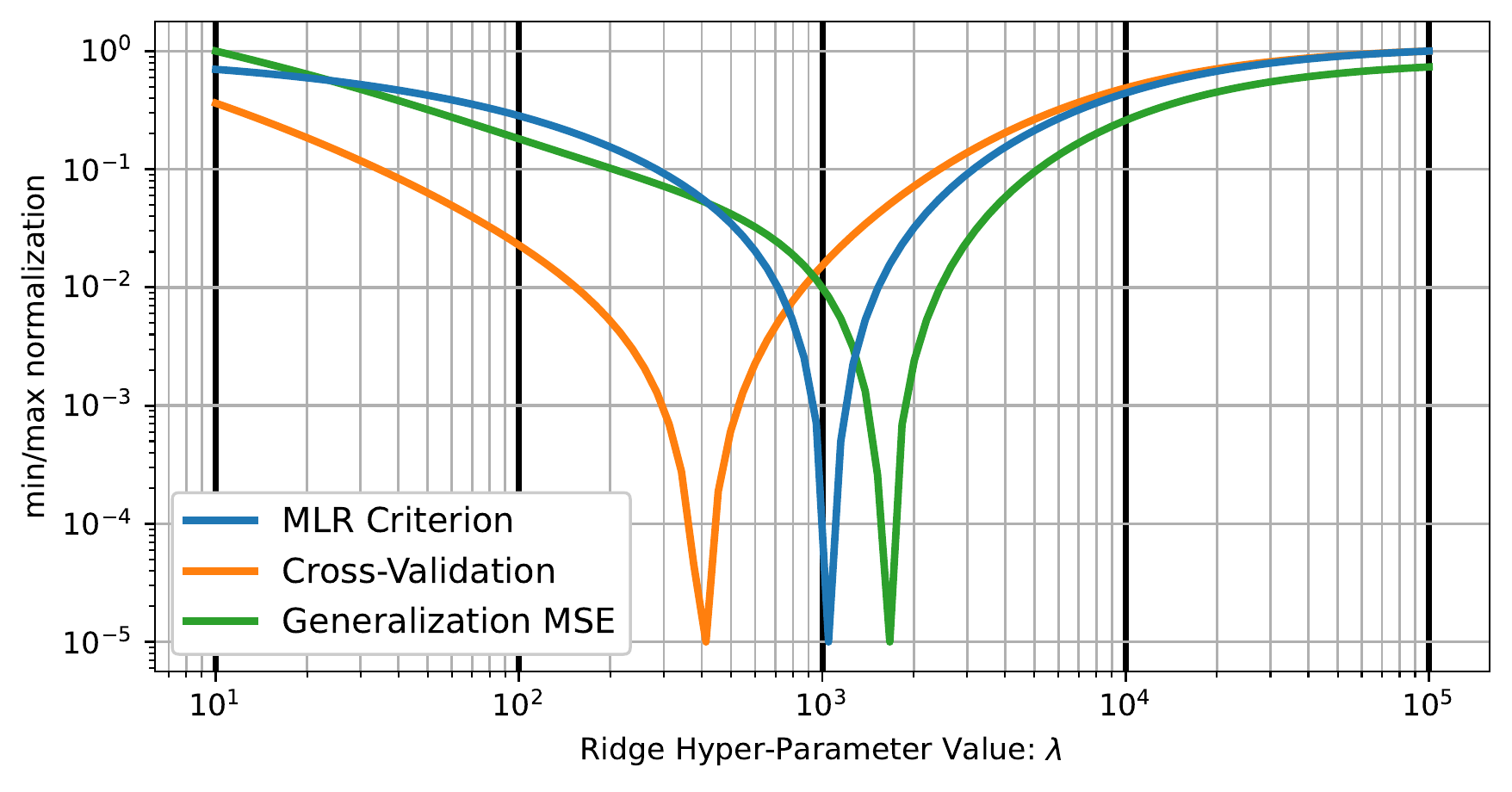}
\caption{Comparison of the \MLR{} criterion (blue), \CV{} criterion ($10$-Fold cross-validation \RMSE{}) (orange) for out-of-sample performance (test set \RMSE) (green) estimation with the Ridge model. We generated synthetic regression data (Appendix \ref{app:syntheticdataMLR}), and train a Ridge model with different levels of regularization $\lambda$. We also train a Ridge model with a randomly permuted target vector. We evaluate the \MLR{} criterion, the $10$-Fold \CV{} \RMSE{} and test \RMSE{} over the $\lambda$ grid and compare their respective argmin, $\widehat{\lambda}_{\MLR}$, $\widehat{\lambda}_{\CV}$ and $\lambda^*$. The goal is to obtain an argmin as close as possible to the optimal one in terms of generalization. Averaging over $100$ seeds, the \RMSE{} test performances with $\widehat{\lambda}_{\MLR}$, $\widehat{\lambda}_{\CV}$ and $\lambda^*$ are $0.7128$, $0.7221$ and $0.7061$ respectively. Above figure is the criterion landscape for random seed $0$. We see that \MLR{} provides a better estimate of the argmin of test \RMSE{} than \CV{}.
}
\label{fig:3pred}
\end{figure}

\textbf{Setting.} Let $\mathcal{D}_{train}=(\bx,\bY)=\{(\bx_i,Y_i)\}_{i=1}^n$ be the $train$-set with $\bx_i\in\R^d$ where $d$ denotes the number of features and $Y_i \in \mathcal{Y}$ where $\mathcal{Y}=\R$ for regression and $\mathcal{Y}$ is a finite set for classification. We optimise the objective $L(\actout(\bm{f}_{\bth}(\bx)),\bY)$ where $\bm{f}_{\bth}(\bx)$ is the output of the last hidden layer, $L$ is the loss function (\MSE{} for regression and \CE{} for classification) and $\actout$ is the activation function ($\Id$ for regression, \sigmoid{} for binary classification and \logsoftmax{} for multiclass).


\textbf{Random permutations.}We build a randomized data set by applying random permutations on the $n$ components of the label vector $\bY$. This randomization scheme presents the advantage of creating an artificial train set $(\bx,\Yperm)$ with marginal distributions of features and labels identical to those in the initial train set but where the connection between features and labels has been removed\footnote{The expected number of fixed points of a permutation drawn uniformly at random is equal to $1$.}. This means that there is no generalizing pattern to learn from the artificial dataset $(\bx,\Yperm)$. We replace the initial loss $L$ by
\begin{align}
   &\MLR(\bth):=L\left(\bY,\actout(\bm{f}_{\bth}(\bx))\right)\notag\\
    &\hspace{2cm}-  L\left(\Yperm,\actout(\bm{f}_{\bth}(\bx))\right).
\label{eq:BLbis1}
\end{align}

The second term on the right-hand side of \eqref{eq:BLbis1} is used to quantify memorization of output layer $\bm{f}_{\bth}$. Indeed, since there is no meaningful pattern linking $\bx$ to $\bY_{\texttt{perm}}$, any $\bm{f}_{\bth}$ which fits $(\bx,\bY_{\texttt{perm}})$ well achieves it via memorization only. We want to rule out such models. By minimizing the \MLR{} loss, we hope to retain only the generalizing patterns.

The \MLR{} approach uses random labels in an original way. In \cite{zhang2016understanding,arpit2017closer}, noise labels are used as a diagnostic tool in numerical experiments. On the theory side, Rademacher Process (RP) is a central tool exploiting random (Rademacher) labels to compute data dependent measures of complexity of function classes used in learning \cite{KoltchinskiiSaintFlour2011}. However, RP are used to derive bounds on the excess risk of already trained models whereas the \MLR{} approach uses randomly permuted labels to train the model.


\noindent
\textbf{Experiment (Fig.\ref{fig:3pred}).} We compare the \MLR{} loss and Cross-Validation (CV) error to the true generalization error in the correlated regression setting described in Appendix \ref{app:syntheticdataMLR}. \MLR{} is a better estimate of the generalization error than \CV, thus yielding a more precise estimate of the optimal hyperparameter $\lambda^*$ than \CV. 

\noindent
\textbf{Theoretical investigation of \MLR{}.} To understand the core mechanism behind the \MLR{} loss, we consider the following toy regression model. Let $\bY = \bx \bm{\beta}^* + \bxi$ with $\bm{\beta}^*\in \R^d$ and isotropic sub-Gaussian noise $\bxi\in\R^n$ ($\mathrm{Cov}(\bxi) = \sigma^2 \I_n$). We consider the class of Ridge models $\mathcal{F}^R = \{f_\lambda(\cdot) = \langle \bm{\beta}_\lambda,\cdot\rangle,\; \lambda>0\}$ with $\bm{\beta}_\lambda =\bm{\beta}_\lambda(\bx,\bY) =(\bx^\top\bx + \lambda \I_d)^{-1}\bx^\top \bY\in\R^d$. Define the risk $R(\lambda) :=  \E_{\bxi}[\|\bx\bm{\beta}^* - \bx\bm{\beta}_\lambda\|_2^2]$, and the optimal parameter 
$\lambda^* = \mathrm{argmin}_{\lambda>0} R(\lambda)$.
We assume for simplicity that $\bx^\top \bx/n$ is an orthogonal projection (denoted $P_{\bx}$) onto a $r$-dimensional subspace of $\R^d$. 
Define the rate
$$
\epsilon_n := \sqrt{\frac{r\sigma^2}{\|\bx\bm{\beta}^*\|_2^2}} + \sqrt{\frac{\|\bx\bm{\beta}^*\|_2^2}{n\sigma^2}}.
$$

\begin{theo}\label{thm1}
Under the above assumptions. If
$ r \sigma^2\ll \|\bx\bm{\beta}^*\|_2^2\ll n \sigma^2$, then we get w.h.p. 
\begin{align*}
\MLR(\lambda)+\|P_{\bx}(\Yperm)\|_2^2
=\left( 1+ o(1)\right)R(\lambda),\quad \forall \lambda >\epsilon_n.
\end{align*}
\end{theo}
Proof is provided in Appendix \ref{app:proofThm1}. In our setting, $\|\bx\bm{\beta}^*\|_2^2/(n\sigma^2)$ is the Signal-to-Noise Ratio \SNR. 
The intermediate \SNR{} regime $ r \sigma^2\ll \|\bx\bm{\beta}^*\|_2^2\ll n \sigma^2$ is the only regime where using Ridge regularization can yield a significant improvement in the prediction. In that regime, the \MLR{} loss can be used to find optimal hyperparameter $\lambda^*$. 
In the high \SNR{} regime $\|\bx\bm{\beta}^*\|_2^2\geq n \sigma^2$, no regularization is needed, i.e. $\lambda^*=0$ is the optimal choice. Conversely in the low \SNR{} regime $  \|\bx\bm{\beta}^*\|_2^2\leq r \sigma^2$, the signal is completely drowned in the noise. Consequently it is better to use the zero estimator, i.e. $\lambda^*=\infty$. \\
In a nutshell, while the high and low \SNR{} regimes correspond to trivial cases where regularization is not useful, in the intermediate regime where regularization is beneficial, \MLR{} is useful. %

\section{The \trainmet{} method to train \DNN{}}
\label{sec:adapcap}

\textbf{The \textbf{\shrinkage{}} operator scheme.}
Consider a \DNN{} architecture with $L$ layers. Denote by $\bth$ the hidden layers weights and by $\bA^{L-1}(\bth, \bigcdot)\,:\, \mathbb{R}^{n\times d} \rightarrow \R^{n\times d_{L-1}}$ the output of the last hidden layer.\\ 
Let $\bla \in \mathbb{R}_+^*$ be the \shrinkage{} parameter and define 
\begin{align}
\label{eq:Pmat}
  &\bW(\bla,\bth,\bx):=\left[\left(\bA^{L-1}\right)^\top\bA^{L-1}+\bla \I\right]^{-1}\left(\bA^{L-1}\right)^\top
\end{align}
where $\bA^{L-1}:=A^{L-1}(\bth, \bx)$ and $\I = \I_{d_{L-1}}$ the identity matrix.
The \shrinkage{} operator is
\begin{align}
\label{eq:Hmat}
  &\bH(\bla,\bth,\bx):=\bA^{L-1}\bW(\bla,\bth,\bx).
\end{align}
During training, the \shrinkage{} operator scheme outputs the following prediction for target vector\footnote{In multiclass setting, replace \bY\, by its one-hot encoding.} $\bY$:
\begin{align}
\label{eq:model2}
\bm{f}_{\bla,\bth,\bY}(\bx)=\bH(\bla,\bth,\bx)\bY,
\end{align}
Note that $(\bla,\bth,\bx, \bY)$ may change at each iteration during training/\GD. To train this \DNN{}, we run a Gradient Descent Optimization scheme over parameters $(\bla,\bth)$
\begin{align}
\label{eq:ridgeMLRNN}
    &(\widehat{\bla}, \widehat{\bth})=
    \underset{\bla>0,\, \bth}{\arg\min} \;
 L\left(\bY,\actout\left(\bm{f}_{\bla, \bth,\bY}(\bx)\right)\right).
\end{align}
Eventually, at test time, we freeze $\bW(\widehat{\bla}, \widehat{\bth}, \bx)$, and obtain our final predictor
\begin{align}
\label{eq:model3}
\bm{f}_{\widehat{\bla}, \widehat{\bth}}(\bigcdot)=\actout\left(A^{L-1}(\widehat{\bth},\bigcdot) \bW(\widehat{\bla}, \widehat{\bth},\bx)\bY\right),
\end{align}
where $\actout$ is the last activation function applied to the output layer. Here, $\bW(\widehat{\bla}, \widehat{\bth},\bx)\bY$ are the weights of the output layer set once and for all using the minibatch ($\bx$, $\bY$) associated with ($\widehat{\bla}, \widehat{\bth}$) in case of batch-learning. Therefore, we recover the architecture of a standard \DNN{} where the output of the hidden layers $A^{L-1}(\widehat{\bth},\bigcdot)$ is multiplied by the weights of the output layer.

\textbf{The \shrinkage{} operator scheme works in a fundamentally different way from \WD}. When we apply the \shrinkage{} operator to the output of the last hidden layer and then use backpropagation to train the \DNN{}, we are indirectly carrying over its $capacity$ control effect to the hidden layers of the \DNN. In other words, we are performing {\it inter-layers} regularization (i.e. regularization across the hidden layers) whereas
\WD{} performs {\it intra-layer} regularization.
We trained a \DNN{} using \WD{} on the one-hand and Tikhonov operator on the other hand while all the other training choices were the same between the two training schemes (same loss $L$, same architecture size, same initialization, same learning rate, etc.). Fig. \ref{fig:1Xcorrelmat} shows that the \shrinkage{} scheme works differently from other $L_2$ regularization schemes like \WD. Indeed, Fig.~ \ref{fig:NewLearningDynamic} reveals that the \shrinkage{} scheme completely changes the learning dynamic during \GD{}.

\noindent
\textbf{Training with \MLR{} loss and the \textbf{\shrinkage{} scheme}.}
We quantify the $capacity$ of our model to memorize labels $\bY$ by $L\left(\bY,\actout\left(\bm{f}_{\bla, \bth,\bY}(\bx)\right)\right)$ w.r.t. to labels $\bY$ where the \shrinkage{} parameter $\bla$ modulates the level the $capacity$ of this model. However, we are not so much interested in adapting the capacity to the train set $(\bx,\bY)$ but rather to the generalization performance on the test set. This is why we replace $L$ by \MLR{} in \eqref{eq:ridgeMLRNN}. Since \MLR{} is a more accurate in-sample estimate of the generalization error than the usual train loss (Theorem \ref{thm1}), we expect \MLR{} to provide better tuning of $\lambda$ and thus some further gain on the generalization performance.\\
Combining \eqref{eq:BLbis1} and \eqref{eq:model2}, we obtain the following train loss of our method.
\begin{align}
   &\MLR(\bla, \bth):=L\biggl(\bY,\actout(\bH(\bla,\bth,\bx)\bY)\biggr)\notag\\
   &\hspace{1cm} -  L\biggl(\Yperm,\actout(\bH(\bla,\bth,\bx)\Yperm)\biggr)
   \label{eq:BLbis1-DNN}
\end{align}
To train this model, we run a Gradient Descent Optimization scheme over parameters $(\bla,\bth)$:
\begin{align}
\label{eq:ridgeMLRNN2}
    &(\widehat{\bla}, \widehat{\bth} )= \mathrm{argmin}_{\bla,\bth|\bla>0}\; 
    \MLR(\bla, \bth).
\end{align}
%
The \trainmet{} predictor is defined again by \eqref{eq:model2} but with weights obtained in \eqref{eq:ridgeMLRNN2} and corresponds to the architecture of a standard \DNN{}. Indeed, at test time, we freeze $\bW(\widehat{\bla},\widehat{\bth},\bx)\bY$ which becomes the weights of the output layer. Once the \DNN{} is trained, the corrupted labels $\Yperm$ and the \shrinkage{} parameter $\widehat{\bla}$ have no further use and are thus discarded. If using Batch-Learning, we use the minibatch $(\bx,\bY)$ corresponding to $(\widehat{\bth},\widehat{\bla})$. In any case, the entire training set can also be discarded once the output layer is frozen.

\textbf{Comments.}\\
$\bullet$ \shrinkage{} is absolutely needed to use \MLR{} on \DNN{} in a differentiable fashion because \FFNN{} have such a high capacity to memorize labels on the hidden layers that the SNR between output layer and target is too high for \MLR{} to be applicable without controlling capacity via the \shrinkage{} operator. Controlling network capacity via HPO over regularization techniques would produce a standard bi-level optimization problem.\\
$\bullet$ The random labels are generated before training and are not updated or changed thereafter. Note that in practice, the random seed used to generate the label permutation has virtually no impact as shown in Table \ref{tab:randomseedimpact}.
\\
$\bullet$ In view of Theorem \ref{thm1}, both terms composing the \MLR{} loss should be equally weighted to produce an accurate estimator of the generalization error.\\
$\bullet$ Note that $\bla$ is not an hyperparameter in \trainmet{} . It is trained alongside $\bth$ by \GD. The initial value $\bla_{\texttt{init}}$ is chosen with a simple heuristic rule. For initial weights $\bth$, we pick the value which maximizes sensitivity of the \MLR{} loss $w.r.t.$ variations of $\bla$ (See \eqref{lambdainiti} in Appendix \ref{sec:appprotocol}).\\
$\bullet$ Both terms of the \MLR{} loss depend on $\bth$ through the quantity $\bH(\bla,\bth,\bx)$, meaning we compute only one derivation graph  $w.r.t.$ $\bH(\bla,\bth,\bx)$.\\
$\bullet$ When using the \textbf{\shrinkage{} operator} during training, we replace a matrix multiplication by a matrix inversion. This operation is differentiable and inexpensive as parallelization schemes provide linear complexity on GPU \cite{SHARMA201331}\footnote{This article states that the time complexity of matrix inversion scales as $J$ as long as $J^2$ threads can be supported by the GPU where $J$ is the size of the matrix.}. Time computation comparisons are provided in Table \ref{tab:runtime}. The overcost depends on the dataset but remains comparable to applying Dropout (\dropout{}) and Batch Norm (\batchnorm{}) on each hidden layers for \DNN{} with depth $3+$.\\
$\bullet$ For large datasets, \trainmet{} can be combined with Batch-Learning. Table \ref{tab:batchsize} in appendix reveals that \trainmet{} works best with large batch-size, but handles very small batches and seeing fewer times each sample much better than regular \DNN{}.

\begin{table*}[h]
\caption{Percentage of experiments where the best performing method belongs to the category (higher is better). For each random train/test split of each considered dataset, we evaluate all methods and than consider two competitions: $\bullet$ without \trainmet{} the \DNN{} category consists of $4$ methods: \MLP, \ResBlock, \SNN{} and \MLP\GLU; $\bullet$ with \trainmet{} the \DNN{} category contains the $4$ following methods instead: \MLP, \trainmet{}\ResBlock, \trainmet{}\SNN{} and \trainmet{}\MLP\GLU. In both competitions, all the other categories contains all the method listed in the benchmark description. For classification \TD{}, we did not report results with \trainmet{} as it under-performs vastly against regular \DNN{} in terms of accuracy and Area Under Curve (AUC), meaning it is not a suitable technique. For regression, \DNN{} compare more favorably when introducing \trainmet{}, especially on the $8$ datasets where the best method obtains a \RMSE{} score under 0.25.}
\label{tab:countsentiredatasets}
\centering
\footnotesize
\begin{tabular}{|c||c|c||c|c||c|c|}
\hline
category&\multicolumn{2}{|c||}{\RMSE{} top~1  on 26 $\TD$}&\multicolumn{2}{|c||}{\RMSE{}  top~1 on 8 \TD}&\multicolumn{2}{|c|}{\binclf{} on 18 \TD }\\
&\multicolumn{2}{|c||}{}&\multicolumn{2}{|c||}{with $\min \RMSE{} < 0.25$}&\multicolumn{2}{|c|}{without \trainmet{}}\\
\hline
&without \trainmet{} &with \trainmet{} &without \trainmet{} &with \trainmet{} &\AUC{} top~1& Err. rate top~1\\
\hline
\GBDT{} & \textbf{39.615} \%& \textbf{36.538} \%& $35.0$ \%& $27.160$ \%& \textbf{61.666} \%& \textbf{73.888}\% \\
\DNN{} & $30.0$ \%& $33.461$ \%& 50.0 \%& \textbf{58.024} \%& $16.666$ \%& $5.5555$\% \\
\RF{} & $18.461$ \%& $18.076$ \%& $15.0$ \%& $14.814$ \%& $15.0$ \%& $10.0$\% \\
\SVM{} & $5.7692$ \%& $6.1538$ \%& $0.0$ \%& $0.0$ \%& $0.0$ \%& $0.0$\% \\
\GLM{} & $4.6153$ \%& $4.6153$ \%& $0.0$ \%& $0.0$ \%& $6.6666$ \%& $7.7777$\% \\
\MARS{} & $1.5384$ \%& $1.1538$ \%& $0.0$ \%& $0.0$ \%& N.A. & N.A.  \\
\CART{} & $0.000$ \%& $0.0$ \%& $0.0$ \%& $0.0$ \%& $0.0$ \%& $2.7777$ \% \\
\hline
\end{tabular}
\end{table*}

\section{Experiments}
\label{sec:exp}




Our goal in this section is to tabulate the impact of \trainmet{} on simple \FFNN{} architectures, on a tabular data benchmark, an ablation and parameter dependence  study, and also a toy few shot learning experiment (Fig. \ref{fig:fewshot}). 

Note that in the main text,
we report only the key results. In supplementary, we provide a detailed description of the benchmarked \FFNN{} architectures and corresponding hyperparameters choices; a dependence study of impact of batchsize, \DO\&\BN, and random seed; the exhaustive results for the tabular benchmark; the implementation choices for compared methods; datasets used with sources and characteristics; datasets preprocessing protocol; hardware implementation. 


\subsection{Implementation details}

Creating a pertinent benchmark for \TD{} is still an ongoing process for ML research community. Because researchers compute budget is limited, arbitrages have to be made between number of datasets, number of methods evaluated, intensity of HPO, dataset size, number of train-test splits. 
We tried to cover a broad set of usecases \cite{paleyes2020challenges} where improving \DNN{} performance compared to other existing methods is relevant, leaving out hours-long training processes relying on HPO to get the optimal performance for each benchmarked method. We detail below how this choice affected the way we designed our benchmark.

\begin{table*}[h]
\caption{Regression task focus: test \RMSE{} (lower is better), P90 (higher is better), and runtime for the $10$ methods from all categories which performed best on $26$ tabular datasets. \RMSE{} is averaged over $10$ train/test splits. The P90 metric measures for each method, the percentage of experiments where the best \RMSE{} is not under 90\% of the method \RMSE{}, meaning it did not underperform too much. \trainmet{} + \SNN{} outperforms the other architectures and \Catboost by a large margin in terms of avg. \RMSE{} but \trainmet{} + \GLU{}\MLP{} performances are more consistent as revealed by the P90 metric, even more so on the $8$ datasets where the best method obtains a \RMSE{} score under $0.25$.}
\label{tab:runtime}
\centering
\footnotesize
\begin{tabular}{|l||c|c|c|c|c|c|}
\hline
method & \RMSE{} avg. & P90  & \RMSE{} avg.  & P90  & avg. runtime  & max  \\
 & all \TD  & all \TD &  \TD with  & \TD with  & avg.  &  runtime \\
 &  &  & $\min\RMSE{} < 0.25$ & $\min\RMSE{} < 0.25$ &  (sec.) &   (sec.) \\
\hline
\hline
\mlrnetselu & \textbf{0.4147} & $55.0$ & $0.1486$ & $32.5$ & \textbf{19.798} & $169.82$ \\
\regularnetglu & $0.4201$ & $54.230$ & $0.1498$ & $40.0$ & $9.8911$ & $35.699$ \\
\mlrnetglu & $0.4206$ & \textbf{60.384} & \textbf{0.1455} & \textbf{56.25} & $22.355$ & $179.88$ \\
\mlrnetresblock & $0.4214$ & $50.0$ & $0.1532$ & $30.0$ & $17.192$ & $166.13$ \\
\Catboost{} & $0.4221$ & $66.538$ & $0.1910$ & $45.0$ & $92.518$ & $315.15$ \\
\regularnetstandard & $0.4230$ & $50.769$ & $0.1601$ & $26.25$ & $4.0581$ & $22.213$ \\
\mlrnetstandard & $0.4233$ & $46.923$ & $0.1566$ & $17.5$ & $17.208$ & $168.10$ \\
\mlrnetfastselu & $0.4245$ & $42.307$ & $0.1591$ & $16.25$ & $7.6670$ & $38.286$ \\
\mlrnetbatchresblock & $0.4257$ & $45.384$ & $0.1580$ & $20.0$ & $194.72$ & \textbf{2654.7} \\
\regularnetselu & $0.4260$ & $40.384$ & $0.1526$ & $18.75$ & \textbf{7.1895} & $32.581$ \\
\hline
\end{tabular}
\end{table*}

\noindent
\textbf{\FFNN{} Architectures.}
For binary classification (\binclf), multiclass classification (\multiclf) and regression (\reg), the output activation/training loss are \sigmoid/\BCE, \logsoftmax/\CE{} and \Id/\RMSE{} respectively. We also implemented the corresponding \MLR{} losses. In all cases, we used the Adam (\Adam) \cite{kingma2014adam} optimizer and the One Cycle Learning Rate Scheduler scheme \cite{,smith2015cyclical}. Early-Stopping is performed using a validation set of size $\min(n*0.2, 2048)$. Unless mentioned otherwise, Batch-Learning is performed with batch size $\bs=\min (n*0.8, 2048)$ and the maximum number of iteration does not depend on the number of epochs and batches per epoch, to cap the training time, in accordance with our benchmark philosophy. We initialized layer weights with Kaiming \cite{kaiming2015}. Then, for \trainmet{}, the \shrinkage{} parameter $\lambda$ is initialized by maximizing the \MLR{} loss sensitivity $w.r.t.$ $\lambda$ on the first mini-batch (See Appendix \ref{sec:appprotocol}). When using \trainmet{}, we used no other additional regularization tricks. Otherwise we used \batchnorm{} and $\dropout = 0.2$ on all hidden layers. Unless mentioned otherwise, we set $\max_{\iter} = 500$ and $\max_{\lr} = 0.01$, hidden layers width $512$ and \ReLu{} activation.


We implemented some architectures detailed in \cite{Klambauer2017,gorishniy2021revisiting}; \MLP: MultiLayer Perceptrons of depth $2$; \ResBlock: Residual Networks with $2$ Resblock of depth $2$; \SNN{} for 
\MLP{} with depth $3$ and \SELU{} activation. We define \GLU{} when hidden layers are replaced with Gated Linear Units. \Fast{} denotes a faster version of \MLP{} and \SNN{} with $\max_{\iter} = 200$ and hidden layers width $256$. \Batch\MLP{} and \Batch\ResBlock{} denote a slower version where the number of epochs is set at $20$ and $50$ respectively and the batch size is set at $min(n,256)$ but the number of iterations is not limited, we enforce a one hour training budget instead. The \Batch{} architectures are outside of the scope of this benchmark and only provided for compute time and performance comparison with iteration bounded versions. In total, we implemented $16$ architectures: \MLP, \Fast\MLP, \Batch\MLP, \SNN, \Fast\SNN, \MLP\GLU, \ResBlock, \Batch\ResBlock; each time trained with and without \trainmet. These where evaluated individually but to count which methods perform best (Table \ref{tab:countsentiredatasets}) we used a restricted set of methods (\#4) for \DNN{}. When the top~1 count is made without \trainmet, we picked \MLP, \ResBlock, \SNN{} and \MLP\GLU. When \trainmet{} is included, we picked \MLP, \trainmet\ResBlock, \trainmet\SNN{} and \trainmet\MLP\GLU. We do so to avoid biasing results in favor of \DNN{} by increasing the number of contenders from this category. See Table~\ref{tab:architectureinfos-app} in the Appendix.

\noindent
\textbf{Other compared methods.} 
We considered \Catboost{} \cite{Prokhorenkova2018}, \XGBoost{} \cite{guestrin2016}, \lightgbm{} \cite{Ke2017} ), \MARS{} \cite{Friedman1991} (py-earth implementation) and the scikit learn implementation of \RF{} and \XRF{} \cite{barandiaran1998random,breiman2001}, Ridge \Kernel{} and \NuSVM \cite{Chihchung2011}, \MLP{} \cite{Hinton89connectionistlearning}, \Enet{} \cite{Zou05}, \Ridge{} \cite{Hoerl1970}, \LAS{} \cite{tibshirani1996}, Logistic regression (\LogReg{} \cite{cox1958}), \texttt{CART}, \texttt{XCART} \cite{Breiman1984,gey2005model,Klusowski2020sparse},\Adaboost and \XGB{} \cite{Breiman1997_arcing,Friedman2001,Friedman2002}. We included a second version of \Catboost{} denoted \Fast\Catboost{}, with hyperparameters chosen to reduce runtime considerably while minimizing performance degradation.

\noindent
\textbf{Benchmarked Tabular Data.} \TD{} are very diverse. We browsed UCI \cite{Dua:2019}, Kaggle and OpenML \cite{OpenML2013}, choosing datasets containing structured columns, $i.i.d.$ samples, one or more specified targets and corresponding to a non trivial learning task, that is the \RF{} performance is neither perfect nor behind the intercept model. We ended up with $44$ datasets (Table~\ref{tab:dataset-app}): UCI $34$, Kaggle $5$ and openml $5$, from medical, marketing, finance, human ressources, credit scoring, house pricing, ecology, physics, chemistry, industry and other domains. Sample size ranges from $57$ to $36584$ and the number of features from $4$ to $1628$, with a diverse range of continuous/categorical mixtures. The tasks include $26$ continuous and ordinal \reg{} and $18$ \binclf{} tasks. Data scarcity is a frequent issue in \TD{} \cite{Chahal2021Small} and Transfer Learning is almost never applicable. However, the small sample regime was not really considered by previous benchmarks. We included $28$ datasets with less than $1000$ samples (\reg{} task:$15$, \binclf{} task:$13$). We also made a focus on the $8$ \reg{} datasets where the smallest \RMSE{} achieved by any method is under $0.25$, this corresponds to datasets where the \SNR is high but the function to approximate is not trivial. For the bagging experiment, we only used the $15$ smallest regression datasets to reduce compute time.

\noindent
\textbf{Dataset preprocessing.}
We applied uniformly the following pipeline: $\bullet$ remove columns with id information, time-stamps, categorical features with more than $12$ modalities (considering missing values as a modality); $\bullet$ remove rows with missing target value; $\bullet$ replace feature missing values with mean-imputation; $\bullet$ standardize feature columns and regression target column. For some regression datasets, we also applied transformations ($e.g.$ $log(\cdot)$ or $\log(1+\cdot)$) on target when relevant/recommended (see Appendix \ref{app:pre-processing}). 

\noindent
\textbf{Training and Evaluation Protocol.}
For each dataset, we used $10$ different train/test splits (with fixed seed for reproducibility) without stratification as it is more realistic. For each dataset and each split, the test set was only used for evaluation and never accessed before prediction. Methods which require a validation set can split the train set only. We evaluated on both train and test set the $\Rdeux$-score and \RMSE{}  for regression and the Accuracy (\Acc) or Area Under Curve \AUC{} for classification (in a {\it one-versus-rest} fashion for multiclass). For each dataset and each we also computed the average performance over the $10$ train/test splits for the following global metrics: PMA, P90, P95 and Friedman Rank. We also counted each time a method outperformed all others (\isTopOne) on one train/test splits of one dataset.

\noindent
\textbf{Meta-Learning and Stacking}
Since the most popular competitors to \DNN{} on \TD{} are ensemble methods, it makes sense to also consider Meta-Learning schemes, as mentionned by \cite{gorishniy2021revisiting}. For a subset of \reg{} datasets, we picked the methods from each category which performed best globally and evaluated bagging models, each comprised of $10$ instances of one unique method, trained with a different seed for the method (but always using the same train/test split), averaging the prediction of the $10$ weak learners. This scheme multiplies training time by $10$, which for most compared methods means a few minutes instead of a few second. Although it has been shown that HPO can drastically increase the performance of some methods on some large datasets, it also most often multiply the compute cost by a factor of $500$ ($5$ Fold \CV $*$ $100$ iterations in \cite{gorishniy2021revisiting}), from several hours to a few days.

\textbf{Benchmark limitations.} This benchmark does not address some interesting but out of scope cases for relevance or compute budget reasons: huge datasets (10M+), specific categorical features handling, HPO, pretraining, Data Augmentation, handling missing values, Fairness, etc., and does not include methods designed for those cases (notably \NODE, \TabNet, FeatureTokenizer, leaving out the comparison/combination of \trainmet{} with these.

\subsection{Tabular Data benchmark results}
\textbf{Main takeaway: \trainmet{} vs regular \DNN{}.}
$\bullet$ Compared  with regular \DNN{}, \trainmet{} is almost irrelevant for classification but almost always improves \reg{} performance. Its impact compounds with the use of \SELU{}, \GLU{} and \ResBlock.\\
$\bullet$ Compute time wise, the overcost of the \shrinkage{} operator matrix inversion is akin to increasing the depth of the network (Table~\ref{tab:runtime}). \\
$\bullet$ There is no SOTA method for \TD. In terms of achieving top~1 performance, \GBDT{} comes first on only less than 40\% of the regression datasets followed by \DNN{} without \trainmet{} at 30\%. Using \trainmet{} to train \DNN{}, the margin between \GBDT{} and \trainmet{}-\DNN{} divides by 3 this gap. (Table~\ref{tab:countsentiredatasets}). In terms of average \RMSE{} performance across all \reg{} datasets, \trainmet{}\SNN{} and \trainmet{}\GLU{}\MLP{} actually comes first before \Catboost{} (Table~\ref{tab:runtime}).\\
$\bullet$ On regression \TD{} where the best achievable \RMSE{} is under $0.25$ \trainmet{} dominates the leaderboard. This confirms our claim that \trainmet{} can delay memorization during training, giving \DNN{} more leeway to capture the most subtle patterns.\\
$\bullet$ Although \trainmet{} reduces the impact of the random seed used for initialization (Table~\ref{tab:randomseedimpact}), it still benefits as much from bagging as other non ensemble methods.

\begin{table}[http!]
\caption{Impact of bagging $10$ instances of the same method, on regression \TD{}. We took the top methods of each category and for $1$ train/test split of $15$ regression \TD{} we trained the method $10$ times with different random seed and averaged the predictions evaluate the potential variation of \RMSE{} with simple method ensembling (lower is better).}
\label{tab:metalearning}
\centering
\footnotesize
\begin{tabular}{|l||c|c|c|}
\hline
top~8 & \RMSE{}  & \RMSE{}  & \RMSE{}  \% \\
best methods & no bag &   bag10 &   variation \% \\
\hline
\hline
\mlrnetselu & $0.3532$ & $0.3322$ & $-5.933$ \\
\mlrnetglu & $0.3482$ & $0.3330$ & $-4.362$ \\
\regularnetselu & $0.3615$ & $0.3374$ & $-6.671$ \\
\regularnetglu & $0.3593$ & $0.3402$ & $-5.296$ \\
\regularnetfast & $0.3698$ & $0.3492$ & $-5.562$ \\
\Catboost{} & $0.3638$ & $0.3610$ & \textit{-0.782} \\
\Fastcat & $0.3879$ & $0.3689$ & $-4.884$ \\
\XRF{} & $0.3813$ & $0.3799$ & $-0.363$ \\
\hline
\end{tabular}
\end{table}





\textbf{Few-shot.} We conducted a toy few-shot learning experiment on MNIST \cite{deng2012mnist} to verify that \trainmet{} is also compatible with \CNN{} architectures in an image multiclass setting. We followed the setting of the pytorch tutorial \cite{mnist2016pytorch} and we repeated the experiment with \trainmet{} but without \DO{} nor \BN{}. The results are detailed in Fig. \ref{fig:fewshot}.

\subsection{Ablation, Learning Dynamic, Dependency study}

Figure~\ref{fig:1Xcorrelmat} shows the impact of both \shrinkage{} and \MLR{} on the trained model. \trainmet{} removes oscillations in learning dynamics Figure~\ref{fig:NewLearningDynamic}. \trainmet{} can handle small batchsize very well whereas standard \MLP{} fails (Table~\ref{tab:batchsize}). \MLP{} trained with \trainmet{} performs better in term of \RMSE{} than when trained with \BN{}+\DO{} (Table~\ref{tab:rmseDOBN}). Combining \trainmet{} with \BN{} or \DO{} does not improve \RMSE{}. The random seed used to generate the label permutation has virtually no impact (Table~\ref{tab:randomseedimpact}).

\section{Conclusion}

We introduced the \MLR{} loss, an in-sample metric for out-of-sample performance, and the \shrinkage{} operator, a training scheme which modulates the capacity of a \FFNN{}. By combining these we obtain \trainmet{}, a training scheme which changes greatly the learning dynamic of \DNN{}. \trainmet{} can be combined advantageously with \CNN, \GLU, \SNN{} and \ResBlock. Its performance are poor on binary classification tabular datasets, but excellent on regression datasets, especially in the high \SNR{} regime were it dominates the leaderboard.\\
Learning on tabular data has witnessed a regain of interest recently. The topic is difficult given the typical data heterogeneity, data scarcity, the diversity of domains and learning tasks and other possible constraints (compute time or memory constraints). We believe that the list of possible topics is so vast that a single benchmark cannot cover them all. It is probably more reasonable to segment the topics and design adapted benchmarks for each.\\ 
In future work, we will investigate development of \trainmet{} for more recent architectures including attention-mechanism to handle heterogeneity in data 
. Finally, we note that the scope of applications for \trainmet{} is not restricted to tabular data. The few experiments we carried out on MNIST and \CNN{} architectures were promising. We shall also further explore this direction in a future work.




\newpage

\bibliographystyle{icml2022}
\bibliography{biblio}

\newpage
\appendix
\onecolumn

\section{The \MLR{} loss}

\subsection{Synthetic data used in Figure \ref{fig:3pred}}
\label{app:syntheticdataMLR}

We generate $n=n_{train}+n_{test}$ i.i.d. observations from the model $(x,Y) \in \R^d\times \R$, $d=80$ $s.t.$ $Y = x^\top \beta^* + \epsilon,$ where $\epsilon\sim\mathcal N(0,\sigma)$, $x\sim N(0_d,\Sigma)$ and $\beta^* \in \R^d$ are mutually independent. We use the following parameters to generate the data:
 $\sigma=100$, $\Sigma = \I_d + H $ with $H_{i,j} = 0.8^{\rho}$ and $\rho \sim \mathcal{U}([1,2])$ and the components of  $\beta^*=\mathbbm{1}_d$. We standardized the observations $\bx_i,Y_i$. We use a train set of size $n_{train}=100$ to compute $\MLR$ and $CV$. We use $n_{test}=1000$ to evaluate the test performance.

    
\subsection{Proof of Theorem \ref{thm1}}
\label{app:proofThm1}

Assume $\mathrm{rank}(\bx)=r$. Consider the SVD of $\frac{1}{\sqrt{n}} \bx$ and denote by $\lambda_1,\ldots, \lambda_r$ the singular values  with corresponding left and right eigenvectors $\{\mathbf{u}_j \}_{j=1}^r \in \R^n$ and $\{\mathbf{v}_j \}_{j=1}^r \in \R^d$:
\begin{align}
    \frac{1}{\sqrt{n}} \bx = \sum_{j=1}^{r} \sqrt{\lambda_j} \mathbf{\bu}_j \otimes \mathbf{\bv}_j.
\end{align}
Define $\bH_\lambda:=\bx(\bx^\top\bx + \lambda \I_d)^{-1}\bx^\top$. We easily get

$$
\bH_\lambda  =\sum_{j=1}^{r} \frac{\lambda_j}{\lambda_j + \lambda/n}  \bu_j \otimes \bu_j.
$$
Compute first the population risk of Ridge model ${\bm{\beta}_\lambda}$. Exploiting the previous display, we have
$$
\|\bx\bm{\beta}^* - \bx \bm{\beta}_{\lambda} \|_2^2 = \sum_{j=1}^r \frac{(\lambda/n)^2}{(\lambda_j + \lambda/n)^2} \langle \bx\bm{\beta}^* , \bu_j\rangle^2 + \sum_{j=1}^r \frac{\lambda_j^2}{(\lambda_j + \lambda/n)^2} \langle \bu_j,\bxi \rangle^2-2\langle (\I_n -\bH_\lambda)\bx\bm{\beta}^*,\bH_\lambda\bxi \rangle.
$$
Taking the expectation $w.r.t.$ $\bxi$ and as $\bxi$ is centered, we get
\begin{align}
\label{eq:truerisk1}
    R(\lambda)= \E_{\xi}\left[\|\bx\bm{\beta}^* - \bx \bm{\beta}_{\lambda} \|_2^2\right] = \sum_{j=1}^r \frac{(\lambda/n)^2}{(\lambda_j + \lambda/n)^2} \langle \bx\bm{\beta}^* , \bu_j\rangle^2 + \sigma^2 \sum_{j=1}^r \frac{\lambda_j^2}{(\lambda_j + \lambda/n)^2}.
\end{align}
Next, compute the representations for the empirical risk
for the original data set $(\bx,\bY)$ and artificial data set $(\bx,\Yperm)$ obtained by random permutation of $\bY$. We obtain respectively
\begin{align}
\label{eq:riskemp1}
 \|\bY - \bx - \bx \bm{\beta}_\lambda(\bx,\bY)\|_2^2 = \|(\I_n -\bH_\lambda)\bY\|_2^2 =\sum_{j=1}^r \left( \frac{\lambda/n}{\lambda_j + \lambda/n} \right)^2 \langle \bY,u_j\rangle^2,\\
\label{eq:riskemp2}
    \|\Yperm - \bx \bm{\beta}_\lambda(\bx,\Yperm)\|_2^2 = \|(\I_n -\bH_\lambda)\Yperm\|_2^2 =\sum_{j=1}^r \left( \frac{\lambda/n}{\lambda_j + \lambda/n} \right)^2 \langle \Yperm,u_j\rangle^2.
\end{align}

As $(\lambda/n)^2=(\lambda_j +\lambda/n)^2-2\lambda_j \lambda/n - \lambda_j^2$, it results the following representation for the \MLR{} loss:
\begin{align}
\label{eq:MLRloss}
 \MLR(\lambda) &=  \|\bY - \bx \bm{\beta}_\lambda(\bx,\bY)\|_2^2- \|\bY - \bx \bm{\beta}_\lambda(\bx,\Yperm)\|_2^2\notag\\
   &= - \|P_{\bx}(\Yperm)\|_2^2 +\sum_{j=1}^{r} \left(\frac{\lambda/n}{\lambda_j +\lambda/n} \right)^{2} \langle  \bY,\bu_j \rangle^2+\sum_{j=1}^{r} \frac{2\lambda_j \lambda/n + \lambda_j^2}{(\lambda_j + \lambda/n)^2} \langle  \Yperm,\bu_j \rangle^2.
\end{align}

Let's consider now the simple case $\lambda_1 = \cdots = \lambda_r = 1$ corresponding to $\bx^\top \bx/n$ being an orthogonal projection of rank $r$. Then, \eqref{eq:truerisk1}  and \eqref{eq:MLRloss} become respectively
\begin{align}
\label{eq:truerisk2}
    R(\lambda)&= \frac{(\lambda/n)^2}{(1 + \lambda/n)^2}  \sum_{j=1}^r \langle \bx\bm{\beta}^* , \bu_j\rangle^2 + \sigma^2 \sum_{j=1}^r\frac{1}{(1 + \lambda/n)^2}=
    \frac{(\lambda/n)^2  }{(1 + \lambda/n)^2}\| \bx\bm{\beta}^*\|^2 +
    \frac{1  }{(1 + \lambda/n)^2}\sigma^2 r.
\end{align}
\begin{align}
\label{eq:MLRloss2}
   \MLR(\lambda)  + \|P_{\bx}(\Yperm)\|_2^2 &= \left(\frac{\lambda/n}{1 +\lambda/n} \right)^{2} \|  P_{\bx}(\bY)\|_2^2+\frac{2 \lambda/n + 1}{(1 + \lambda/n)^2} \| P_{\bx}(\Yperm)\|_2^2.
\end{align}
Since $\|P_{\bx}(\Yperm)\|_2^2$ does not depend on $\lambda$, minimizing $\MLR(\lambda)$ is equivalent to minimizing
\begin{align}
\label{eq:MLRequiv}
    \left(\frac{\lambda/n}{1 +\lambda/n} \right)^{2} \|  P_{\bx}(\bY)\|_2^2+\frac{2 \lambda/n + 1}{(1 + \lambda/n)^2} \| P_{\bx}(\Yperm)\|_2^2.
\end{align}

Comparing the previous display with \eqref{eq:truerisk1}, we observe that $\MLR(\lambda)+ \|P_{\bx}(\Yperm)\|_2^2$ looks like the population risk $\E_{\xi}\left[\|\bx\bm{\beta}^* - \bx \bm{\beta}_{\lambda} \|_2^2\right]$.


Next state the following fact proved in section~\ref{sec:ProofPostulat}
\begin{postulat}
\label{eq:postulat1}
\begin{align}
\label{eq:KaMLRloss2}
      \MLR(\lambda)  + \|P_{\bx}(\Yperm)\|_2^2 &= \left(\frac{1+\lambda/n+\avoir}{1 +\lambda/n} \right)^{2} \widehat R(\lambda+n\avoir)-\frac{\avoir\| P_{\bx}(\Yperm)\|_2^2}{(1 +\lambda/n)^2}
\end{align}

\end{postulat}

Before proving our theorem, define the following quantity $\avoir$ 
\begin{align*}
    \avoir:=\frac{\|P_{\bx}(\Yperm)\|_2^2}{\|P_{\bx}(\bY)\|_2^2}  
\end{align*} 
and state an intermediate result (lemma~\ref{eq:lemma2}- proved in section~\ref{sec:ProofLemme2})
\bigskip

\begin{lemma}
\label{eq:lemma2}
Consider the simple case $\lambda_1 = \cdots = \lambda_r = 1$ corresponding to $\bx^\top \bx/n$ being an orthogonal projection of rank $r$ and define 
\begin{align}
\label{eq:empiriquerisk2}
    \widehat R(\lambda)&=
    \frac{(\lambda/n)^2  }{(1 + \lambda/n)^2}\|P_{\bx}(\bY)\|_2^2 +
    \frac{1  }{(1 + \lambda/n)^2}\|P_{\bx}(\Yperm)\|_2^2.
\end{align}
Then, in the intermediate \SNR{} regime $r \sigma^2\ll \|\bx\bm{\beta}^*\|_2^2\ll n \sigma^2$, it comes 
\begin{eqnarray}
\label{eq:diffDesRisque}
    \widehat R(\lambda+n\avoir)&=&\widehat R(\lambda)\left(1+o\left(\frac{a}{(\lambda/n)}\right)\right)\quad \textit{w.h.p.},\\
\label{eq:Deltadesrisque}
  \widehat R(\lambda))&=&R(\lambda)\left(1+O\left(\epsilon_n\right)\right)\quad \textit{w.h.p.}
\end{eqnarray}
where $\epsilon_n=\sqrt{\frac{\|\bx\bm{\beta}^*\|_2^2}{n\sigma^2} }+\sqrt{\frac{r \sigma^2}{\|\bx\bm{\beta}^*\|_2^2 }}$.
\end{lemma}

Starting from \eqref{eq:KaMLRloss2} and using successively \eqref{eq:diffDesRisque} and \eqref{eq:Deltadesrisque}, we have \textit{w.h.p.}

\begin{eqnarray*}
\MLR(\lambda)  + \|P_{\bx}(\Yperm)\|_2^2 &=& \left(1+\frac{\avoir}{1 +\lambda/n} \right)^{2} \widehat R(\lambda+n\avoir)-\frac{\avoir\| P_{\bx}(\Yperm)\|_2^2}{(1 +\lambda/n)^2}
\end{eqnarray*}

Combining \eqref{eq:diffDesRisque},   \eqref{eq:Deltadesrisque} and \eqref{eq:KaMLRloss2}

\begin{eqnarray*}
\MLR(\lambda)  + \|P_{\bx}(\Yperm)\|_2^2 &=& \left(1+\frac{\avoir}{1 +\lambda/n} \right)^{2}\widehat R(\lambda)\left(1+o\left(\frac{a}{(\lambda/n)}\right)\right)-\frac{\avoir\| P_{\bx}(\Yperm)\|_2^2}{(1 +\lambda/n)^2}\\
&=&\left(1+\frac{\avoir}{1 +\lambda/n} \right)^{2}R(\lambda)\left(1+O(\epsilon_n)+o\left(\frac{a}{(\lambda/n)}\right)\right)-\frac{\avoir\| P_{\bx}(\Yperm)\|_2^2}{(1 +\lambda/n)^2}
\end{eqnarray*}

Moreover by \eqref{eq:KPermpiY} in lemma~\ref{lemma:1}, we have
\begin{align*}
    \|P_{\bx}(\Yperm)\|_2^2& = r\sigma^2\left(1 + O\left(\sqrt{\frac{\|\bx\bm{\beta}^*\|^2_2}{n\sigma^2}} \right)\right).
\end{align*}

Then, we get

\begin{eqnarray*}
\MLR(\lambda)  + \|P_{\bx}(\Yperm)\|_2^2 &=&\left(1+\frac{\avoir}{1 +\lambda/n} \right)^{2}R(\lambda)\left(1+O(\epsilon_n)\right)-\frac{\avoir r\sigma^2}{(1 +\lambda/n)^2} \left(1 + O\left(\sqrt{\frac{\|\bx\bm{\beta}^*\|^2_2}{n\sigma^2}} \right)\right).
\end{eqnarray*}

In the intermediate \SNR{} regime $r \sigma^2\ll \|\bx\bm{\beta}^*\|_2^2\ll n \sigma^2$, by lemma~\ref{lemma:1} (section~\ref{sec:ProofLemme1})
\begin{align}
\label{eq:SNRnoise1}
    \avoir:=\frac{\|P_{\bx}(\Yperm)\|_2^2}{\|P_{\bx}(\bY)\|_2^2}  = \frac{\sigma^2 r}{\|\bx\bm{\beta}^*\|_2^2}(1 + o(1)),\quad\text{\textit{w.h.p.}}.
\end{align} 

Then, $r \sigma^2\ll \|\bx\bm{\beta}^*\|_2^2\ll n \sigma^2$ $\Rightarrow$ $\epsilon_n=o(1)$ and $\forall\lambda/n>>\avoir$
\begin{align*}
    \MLR(\lambda) + \|P_{\bx}(\Yperm)\|_2^2 &=  R(\lambda)(1+o(1)). 
\end{align*}

\subsection{Proof of Lemma~\ref{eq:lemma2}}
\label{sec:ProofLemme2}

Consider the simple case $\lambda_1 = \cdots = \lambda_r = 1$ corresponding to $\bx^\top \bx/n$ being an orthogonal projection of rank $r$. Recall

\begin{align}
\label{eq:2empiriquerisk2}
    \widehat R(\lambda)&=
    \frac{(\lambda/n)^2  }{(1 + \lambda/n)^2}\|P_{\bx}(\bY)\|_2^2 +
    \frac{1  }{(1 + \lambda/n)^2}\|P_{\bx}(\Yperm)\|_2^2.
\end{align}

\paragraph{Proof of equation \eqref{eq:diffDesRisque}.}
We have
\begin{eqnarray*}
    \widehat R(\lambda+n\avoir)&=&
    \|P_{\bx}(\bY)\|_2^2\frac{(\lambda/n+\avoir)^2  }{(1 + \lambda/n+\avoir)^2} 
  +  \|P_{\bx}(\Yperm)\|_2^2\frac{1  }{(1 + \lambda/n+\avoir)^2}
\end{eqnarray*}

Next in the intermediate \SNR{} regime $r \sigma^2\ll \|\bx\bm{\beta}^*\|_2^2\ll n \sigma^2$, by lemma~\ref{lemma:1} (section~\ref{sec:ProofLemme1})
\begin{align*}
    \avoir:=\frac{\|P_{\bx}(\Yperm)\|_2^2}{\|P_{\bx}(\bY)\|_2^2}  = \frac{\sigma^2 r}{\|\bx\bm{\beta}^*\|_2^2}(1 + o(1)),\quad\text{\textit{w.h.p.}}.
\end{align*}

Then, $\forall (\lambda/n)>>\avoir$, we get
\begin{eqnarray*}
\frac{(\lambda/n+\avoir)^2  }{(1 + \lambda/n+\avoir)^2}&=&\frac{(\lambda/n)^2}{(1+(\lambda/n))^2}\left[\frac{1+\frac{2a}{(\lambda/n)}+\left(\frac{2a}{(\lambda/n)}\right)^2}{(1+\frac{a}{1+(\lambda/n)})^2}\right]=\frac{(\lambda/n)^2}{(1+(\lambda/n))^2}(1+o\left(\frac{a}{(\lambda/n)}\right),\\
\frac{1 }{(1 + \lambda/n+\avoir)^2}&=&\frac{1}{(1+(\lambda/n))^2}\left[\frac{1}{(1+\frac{a}{1+(\lambda/n)})^2}\right]=\frac{1}{(1+(\lambda/n))^2}(1+o\left(\frac{a}{(\lambda/n)}\right).
\end{eqnarray*}
Therefore, we get the first ingredient to prove our theorem:
\begin{eqnarray*}
    \widehat R(\lambda+n\avoir)&=&\widehat R(\lambda)(1+o\left(\frac{a}{(\lambda/n)}\right).
\end{eqnarray*}

\paragraph{Proof of equation \eqref{eq:Deltadesrisque}.}
Compute now
\begin{eqnarray*}
  \widehat R(\lambda)-  R(\lambda)&=&\frac{(\lambda/n)^2  }{(1 + \lambda/n)^2}\| \left(\|P_{\bx}((\bY)\|_2^2-\bx\bm{\beta}^*\|^2\right) +
    \frac{1  }{(1 + \lambda/n)^2}\left(\|P_{\bx}(\Yperm)\|_2^2-\sigma^2 r\right).
\end{eqnarray*}
As by lemma~\ref{lemma:1}, we have \textit{w.h.p.}
\begin{align}
\label{eq:kKSNRnoise1}
    \|P_{\bx}(\bY)\|_2^2 &=\|\bx\bm{\beta}^*\|_2^2 \left( 1+ O\left( \sqrt{\frac{r \sigma^2}{\|\bx\bm{\beta}^*\|_2^2 }}
    \right) \right),\\
    \label{eq:kKPermpiY}
    \|P_{\bx}(\Yperm)\|_2^2& = r\sigma^2\left(1 + O\left(
    \sqrt{\frac{\|\bx\bm{\beta}^*\|_2^2}{n\sigma^2} }\right)\right).
\end{align}

Then,  we get the second ingredient to prove our theorem. 
\begin{eqnarray*}
  \widehat R(\lambda)&=& R(\lambda)\left(1+\left(O\left( \sqrt{\frac{r \sigma^2}{\|\bx\bm{\beta}^*\|_2^2 }}
    \right) + O\left(
    \sqrt{\frac{\|\bx\bm{\beta}^*\|_2^2}{n\sigma^2} }\right)
  \right)\right)=R(\lambda)\left(1+O\left(\epsilon_n\right)\right),
\end{eqnarray*}
where  $$\epsilon_n:= \sqrt{\frac{r \sigma^2}{\|\bx\bm{\beta}^*\|_2^2 }}
    +
    \sqrt{\frac{\|\bx\bm{\beta}^*\|_2^2}{n\sigma^2} }.$$

\subsection{Fact~\ref{eq:postulat1}}
\label{sec:ProofPostulat}

Compute the following quantity
\begin{eqnarray*}
I&:=&\left(\frac{1+\lambda/n+\avoir}{1 +\lambda/n} \right)^{2} \widehat R(\lambda+n\avoir)-\frac{\avoir\| P_{\bx}(\Yperm)\|_2^2}{(1 +\lambda/n)^2}\\
&=&\frac{(\lambda/n+\avoir)^2  }{(1 +\lambda/n)^2}\|P_{\bx}(\bY)\|_2^2
    +
    \frac{1  }{(1 +\lambda/n)^2}\|P_{\bx}(\Yperm)\|_2^2
  -\frac{\avoir\| P_{\bx}(\Yperm)\|_2^2}{(1 +\lambda/n)^2}\\
 &=&
    \frac{(\lambda/n)^2  }{(1 +\lambda/n)^2}\|P_{\bx}(\bY)\|_2^2  +
    \frac{\avoir^2  }{(1 +\lambda/n)^2}\|P_{\bx}(\bY)\|_2^2
    +
    \frac{2(\lambda/n)\avoir  }{(1 +\lambda/n)^2}\|P_{\bx}(\bY)\|_2^2\\
&&+
    \frac{1  }{(1 +\lambda/n)^2}\|P_{\bx}(\Yperm)\|_2^2
        -\frac{\avoir\| P_{\bx}(\Yperm)\|_2^2}{(1 +\lambda/n)^2}.
\end{eqnarray*}
By \eqref{eq:SNRnoise1} we have $\avoir\|P_{\bx}(\bY)\|_2^2=\| P_{\bx}(\Yperm)\|_2^2$, then
\begin{eqnarray*}
I&=&
    \frac{(\lambda/n)^2  }{(1 +\lambda/n)^2}\|P_{\bx}(\bY)\|_2^2  +
    \frac{\avoir  }{(1 +\lambda/n)^2}\| P_{\bx}(\Yperm)\|_2^2
    +
    \frac{2(\lambda/n)  }{(1 +\lambda/n)^2}\| P_{\bx}(\Yperm)\|_2^2\\
&&+\frac{1  }{(1 +\lambda/n)^2}\|P_{\bx}(\Yperm)\|_2^2
        -\frac{\avoir\| P_{\bx}(\Yperm)\|_2^2}{(1 +\lambda/n)^2}\\
        &=& \frac{(\lambda/n)^2  }{(1 +\lambda/n)^2}\|P_{\bx}(\bY)\|_2^2  
    +
    \frac{2(\lambda/n) +1 }{(1 +\lambda/n)^2}\| P_{\bx}(\Yperm)\|_2^2\\
    &=& \MLR(\lambda)  + \|P_{\bx}(\Yperm)\|_2^2. 
\end{eqnarray*}

Therefore, we get 
\begin{align*}
      \MLR(\lambda)  + \|P_{\bx}(\Yperm)\|_2^2 &= \left(\frac{1+\lambda/n+\avoir}{1 +\lambda/n} \right)^{2} \widehat R(\lambda+n\avoir)-\frac{\avoir\| P_{\bx}(\Yperm)\|_2^2}{(1 +\lambda/n)^2}.
\end{align*}

\subsection{Lemma~\ref{lemma:1}}
\label{sec:ProofLemme1}

\begin{lemma}
\label{lemma:1}
Under the assumption considered in Theorem~\ref{thm1} and in the intermediate \SNR{} regime $ r \sigma^2\ll \|\bx\bm{\beta}^*\|_2^2\ll n \sigma^2$, we have \textit{w.h.p.}
\begin{align}
\label{eq:Knorm1}
    \|P_{\bx}(\bY)\|_2^2 &=\|\bx\bm{\beta}^*\|_2^2 \left( 1+ O\left( \sqrt{\frac{r \sigma^2}{\|\bx\bm{\beta}^*\|_2^2 }}
    \right) \right),\\
    \label{eq:KPermpiY}
    \|P_{\bx}(\Yperm)\|_2^2& = r\sigma^2\left(1 + O\left(\sqrt{\frac{\|\bx\bm{\beta}^*\|^2_2}{n\sigma^2}} \right)\right).
\end{align}
Therefore,
\begin{align}
\label{eq:KSNRnoise1}
    \frac{\|P_{\bx}(\Yperm)\|_2^2}{\|P_{\bx}(\bY)\|_2^2}  &= \frac{\sigma^2 r}{\|\bx\bm{\beta}^*\|_2^2}(1 + o(1)).
\end{align}
\end{lemma}

\paragraph{Proof of \ref{eq:Knorm1}.}
Since $\bxi$ is subGaussian, we have $\|P_{\bx}(\bxi)\|_2^2 = r \sigma^2 +O\left(\sqrt{r \sigma^2 }\right)$ \textit{w.h.p.} \cite{Massart2000}. Thus, under the \SNR{} condition ($\|\bx\bm{\beta}^*\|_2^2 \gg r \sigma^2$), we get
$$
\frac{\|P_{\bx}(\bxi)\|_2^2}{\|\bx\bm{\beta}^*\|_2^2 } \lesssim \frac{r \sigma^2 }{\|\bx\bm{\beta}^*\|_2^2} \ll 1,\quad \text{\textit{w.h.p.}}
$$
Consequently and as $\langle \bx\bm{\beta}^*,\bxi \rangle$ is subGaussian, it comes
\begin{align}
\label{eq:norm1}
    \|P_{\bx}(\bY)\|_2^2 &= \|\bx\bm{\beta}^*\|_2^2 + 2\langle \bx\bm{\beta}^*,\bxi \rangle + \|P_{\bx}(\bxi)\|_2^2\notag\\
    &=\|\bx\bm{\beta}^*\|_2^2 \left( 1+ 2\frac{\langle \bx\bm{\beta}^*,\bxi \rangle}{\|\bx\bm{\beta}^*\|_2^2 } + \frac{\|P_{\bx}(\bxi)\|_2^2}{\|\bx\bm{\beta}^*\|_2^2 }\right)\notag\\
    &=\|\bx\bm{\beta}^*\|_2^2 \left( 1+ O\left( \sqrt{\frac{r \sigma^2}{\|\bx\bm{\beta}^*\|_2^2 }}
    \right) \right)\quad \text{\textit{w.h.p.}}.
\end{align}

\paragraph{Proof of \ref{eq:KPermpiY}.}
Next we recall that $\Yperm = \pi(\bY) = \pi(\bx\bm{\beta}^*) + \pi(\bxi)$ for some $\pi$ drawn uniformly at random in the set of permutation of $n$ elements.  Then
$$
\|P_{\bx}(\Yperm)\|_2^2=\|P_{\bx}(\pi(\bY))\|_2^2 = \|P_{\bx}(\pi(\bx\bm{\beta}^*))\|_2^2 + 2\langle P_{\bx}(\pi(\bx\bm{\beta}^*)), P_{\bx}(\pi(\bxi)) \rangle  + \|P_{\bx}(\pi(\bxi)) \|_2^2.
$$


\begin{itemize}
    \item Exploiting again subGaussianity of $\bxi$, we get $\|P_{\bx}(\pi(\bxi)) \|_2^2 = r \sigma^2 + O(\sqrt{r\sigma^2})$ \textit{w.h.p.}.
    \item We recall that the n-dimensional vector $\bx\bm{\beta}^*$ lives in a $r$-dimensional subspace of $\R^n$ with $r\ll n$. Consequently, randomly permuting its components creates a new vector $\pi(\bx\bm{\beta}^*)$ which is almost orthogonal to $\bx\bm{\beta}^*$:
$$
\frac{\langle \pi(\bx\bm{\beta}^*), \bx\bm{\beta}^* \rangle }{\|\bx\bm{\beta}^*\|_2^2}\lesssim \frac{r}{n}\ll 1,\quad\text{and}\quad \|P_{\bx}(\pi(\bx\bm{\beta}^*))\|_2^2 \lesssim \frac{r}{n} \|\bx\bm{\beta}^*\|_2^2,\quad \text{\textit{w.h.p.}}.
$$
\item In the non trivial \SNR{} regime $r \sigma^2\ll \|\bx\bm{\beta}^*\|_2^2\ll n \sigma^2$ where Ridge regularization is useful, the dominating term in the previous display is $\|P_{\bx}(\pi(\bxi)) \|_2^2$.
\end{itemize}

\paragraph{Proof of \ref{eq:KSNRnoise1}.}
Consequently,  we get 
\begin{align}
    \|P_{\bx}(\pi(\bY))\|_2^2 = r\sigma^2\left(1 + O\left(\sqrt{\frac{\|\bx\bm{\beta}^*\|^2_2}{n\sigma^2}} \right)\right)\quad \text{\textit{w.h.p.}}.\qquad\qquad\qquad\qquad\qquad\square
\end{align}

\newpage
\clearpage

\begin{table}[h]
\caption{Impact of \DO{} and \BN{} on \trainmet\MLP\Fast{} on ConcreteSlump. \DO{} and  \BN{} are applied uniformly on all hidden layers. \trainmet interacts well with \SeLU{}, \GLU{}, \ResBlock{} but not with \DO{} or \BN{}.}
\label{tab:rmseDOBN}
\centering
\footnotesize
\begin{tabular}{|l||c|c|}
\hline
dropout & \RMSE{} without \BN & \RMSE{} with \BN \\
\hline
\hline
$0.0$ & $0.1625$ & $0.1872$ \\
$0.1$ & $0.1850$ & $0.1847$ \\
$0.2$ & $0.1919$ & $0.1954$ \\
$0.3$ & $0.2055$ & $0.2067$ \\
$0.4$ & $0.2176$ & $0.2202$ \\
$0.5$ & $0.2342$ & $0.2352$ \\
\hline
\end{tabular}
\end{table}

\begin{table}[h]
\caption{Impact of batch-size on \trainmet{}\MLP{}\Fast{} and \MLP{}\Fast{} on the  \texttt{Abalone} dataset $(n=3341)$. Since the number of iterations is bounded $(max iter = 200)$, the number of time each sample is seen during training diminishes when batch sizes diminishes in our experiments. \trainmet{} increases the resilience to issues caused by very small batch sizes under fixed number of iterations constraints.}
\label{tab:batchsize}
\centering
\footnotesize
\begin{tabular}{|l||c|c|}
\hline
batchsize & \MLP\Fast & \trainmet{}\Fast \\
\hline
\hline
$16$ & $68.646$ & $0.6756$ \\
$32$ & $7.0196$ & $0.6733$ \\
$64$ & $2.1226$ & $0.6658$ \\
$256$ & $0.6644$ & $0.6573$ \\
$512$ & $0.6626$ & $0.6547$ \\
$1024$ & $0.6620$ & $0.6522$ \\
$2048$ & $0.6596$ & $0.6522$ \\
\hline
\end{tabular}
\end{table}

\begin{table}[h]
\caption{Random Seed Impact: $1000$ method seeds on $1$ train/test split for ConcreteSlump, the only source of randomness for \MLP\Fast{} and \shrinkage\MLP\Fast{} is the weight initialization. Strikingly, \shrinkage does not just improve performance but also reduces the standard deviation, meaning the impact of initial weights on final results. \trainmet{}\MLP\Fast{} $T=1$ uses the \MLR{} loss given in \eqref{eq:BLbis1-DNN}, which introduces randomness when generating label permutations. In all other experiments presented in this article, we use $T = 16$ seeds for generating label permutations and average the loss over these $16$ label vectors, which is enough to offset the randomness introduced with \MLR{}.}
\label{tab:randomseedimpact}
\centering
\footnotesize
\begin{tabular}{|l||c|c|}
\hline
\DNN & \RMSE{} avg. & \RMSE{} std. \\
\hline
\hline
\MLP\Fast & $0.1675$ & $0.0148$ \\
\shrinkage\MLP\Fast & $0.0930$ & $0.0071$ \\
\trainmet{}\MLP\Fast{} $T=1$ & $0.0944$ & $0.0091$ \\
\trainmet{}\MLP\Fast{} $T=16$ & $0.0918$ & $0.0084$ \\
\hline
\end{tabular}
\end{table}


\section{Training protocol}
\label{sec:appprotocol}
\paragraph{Initialization of the \shrinkage{} parameter.} 

We select $\bla_{init}$ which maximizes sensitivity of the $\MLR{}$ objective to variation of $\bla$. In practice, the following heuristic proved successful on a wide variety of data sets. We pick $\bla_{init}$ by running a grid-search on the finite difference approximation for the derivative of \MLR{} in (\ref{lambdainiti}) on the grid 
$\cG_{\bla}=\left\lbrace \bla^{(k)} = 10^{-1}\times10^{5 \times k / 11}\;: \;   k = 0, \cdots, 11 \right\rbrace$:
\begin{eqnarray}
\label{lambdainiti}
\bla_{init}=\sqrt{\bla^{(\hat {k}})\,\bla^{(\hat {k}+1)}},
\end{eqnarray}
where
\begin{eqnarray*}
\hat {k}=\arg\max\left\{ \left(\MLR(\bla^{(k+1)},\bth) - \MLR(\bla^{{(k)}},\bth)\right),\, \bla^{(k)}\in\cG_{\bla}\right\}.
\end{eqnarray*}
Our empirical investigations revealed that this heuristic choice is close to the optimal oracle choice of $\bla_{init}$ on the test set. 

From a computational point of view, the overcost of this step is marginal because we only compute he SVD of $\bA^{L-1}$ once and we do not compute the derivation graph of the $11$ matrix inversions or of the unique forward pass. We recall indeed that the \shrinkage{} parameter $\bla$ is \textbf{not an hyperparameter} of our method; it is trained alongside the weights of the \DNN{} architecture. 

\paragraph{Comments.} We use the generic value ($T=16$) for the number of random permutations in the computation of the \MLR{} loss. This choice yields consistently good results overall. This choice of permutations has little impact on the value of the \MLR{} loss. In addition, when $T=16$, GPU parallelization is still preserved.


\section{Further discussion of existing works comparing \DNN{} to other models on \TD}
\label{app:HPO-bib}

We complement here our discussion of existing benchmarks in the introduction. \DL{} are often beaten by other types of algorithms on tabular data learning tasks. Very recently, there has been a renewed interest in the subject, with several new methods. See \cite{borisov2021deep} for an extensive review of the state of the art on tabular datasets.


The comparison between simple \DNN{}, \NODE, \TabNet, \RF{} and \GBDT{} on \TD{} was made concomitantly by \cite{kadra2021welltuned}, \cite{SHWARTZZIV202284} and \cite{gorishniy2021revisiting}. Their benchmark are more oriented towards an \AutoML{} approach than ours, as they all use heavy HPO, and report training times in minutes/hours, even for some small and medium size datasets.
 
The benchmark in \cite{kadra2021welltuned} compares \FFNN{} to \GBDT{}, \NODE, \TabNet, ASK-\GBDT{} (Auto-sklearn) also using heavy HPO, on $40$ \TD{} (with size ranging from $n=452$ to $416k+$). They reported that, after 30 minutes of HPO time, regularization cocktails for \MLP{} are statistically significantly better than XGBoost (See Table 3 there). We did not find a similar experiment for \Catboost{}.

The benchmark in \cite{SHWARTZZIV202284} includes $11$ datasets from OpenML, Kaggle, Pascal, and MSLR, Million song with $n$ ranging from $7k$ to $1M+$. They compared \XGBoost, \NODE, \TabNet, 1D-CNN, DNF-Net \cite{katzir2021netdnf} and ensemble of these methods.

These 2 benchmarks give mixed signals on how \DNN{} compare to other methods with \cite{kadra2021welltuned} being more optimistic than \cite{SHWARTZZIV202284}.

\section{Benchmark description}

\subsection{Our Benchmark philosophy}


Current benchmarks usually heavily focus on the AutoML (\cite{SHWARTZZIV202284,zoller2021benchmark,yao2018taking,he2021automl}) usecase \cite{zimmer-tpami21a, feurer2020auto}, using costly HPO over a small collection of popular large size datasets, which raised some concerns \cite{koch2021reduced,denton2021genealogy}. 

Creating a pertinent benchmark for Tabular Data (\TD{}) is still an ongoing process for ML research community. Because researchers compute budget is limited, arbitrages have to be made between number of datasets, number of methods evaluated, intensity of HPO, dataset size, number of train-test splits. 
We tried to cover a broad set of usecases where improving \DNN{} performance compared to other existing methods is relevant, leaving out hours-long training processes relying on HPO to get the optimal performance for each benchmarked method.

\subsection{Datasets/preprocessing}

\subsubsection{Datasets} 

Our benchmark includes 44 tabular datasets with 26 regression and 18 classification tasks. Table \ref{tab:dataset-app} contains the exhaustive description of the datasets included in our benchmark.

\begin{table}[http!]
\caption{Datasets description}
\label{tab:dataset-app}
\centering
\tiny
\begin{tabular}{|l||l|c|r|c|c|c|c|c|c|}
\hline
id & name & task & target & n & p & \#cont. & \#cat. & bag  & $min\RMSE$ \\
 &  &  &  index &  &  &  &  & exp. & $<0.25$ \\
\hline
\hline
$0$ & Cervical Cancer Behavior Risk & C & $-1$ & $57$ & $149$ & $19$ & $14$ &  &  \\
$1$ & Concrete Slump Test & R & $-1$ & $82$ & $9$ & $9$ & $0$ & \checkmark & \checkmark \\
$2$ & Concrete Slump Test & R & $-2$ & $82$ & $9$ & $9$ & $0$ & \checkmark & \checkmark \\
$3$ & Concrete Slump Test & R & $-3$ & $82$ & $9$ & $9$ & $0$ & \checkmark & \checkmark \\
$4$ & Breast Cancer Coimbra & C & $-1$ & $92$ & $9$ & $9$ & $0$ &  &  \\
$5$ & Algerian Forest Fires Dataset Sidi-Bel Abbes & C & $-1$ & $96$ & $14$ & $10$ & $1$ &  &  \\
$6$ & Algerian Forest Fires Dataset Bejaia & C & $-1$ & $97$ & $16$ & $12$ & $1$ &  &  \\
$7$ & restaurant-revenue-prediction & R & $-1$ & $109$ & $330$ & $37$ & $39$ &  &  \\
$8$ & Servo & R & $-1$ & $133$ & $21$ & $2$ & $4$ & \checkmark & \checkmark \\
$9$ & Computer Hardware & R & $-1$ & $167$ & $7$ & $7$ & $0$ & \checkmark & \checkmark \\
$10$ & Breast Cancer & C & $0$ & $228$ & $42$ & $1$ & $9$ &  &  \\
$11$ & Heart failure clinical records & C & $-1$ & $239$ & $12$ & $7$ & $5$ &  &  \\
$12$ & Yacht Hydrodynamics & R & $-1$ & $245$ & $22$ & $4$ & $2$ & \checkmark & \checkmark \\
$13$ & Ionosphere & C & $-1$ & $280$ & $33$ & $32$ & $1$ &  &  \\
$14$ & Congressional Voting Records & C & $0$ & $348$ & $48$ & $0$ & $16$ &  &  \\
$15$ & Cylinder Bands & C & $-1$ & $432$ & $111$ & $1$ & $19$ &  &  \\
$16$ & QSAR aquatic toxicity & R & $-1$ & $436$ & $34$ & $8$ & $3$ & \checkmark &  \\
$17$ & Optical Interconnection Network  & R & $7$ & $512$ & $26$ & $6$ & $5$ & \checkmark & \checkmark \\
$18$ & Optical Interconnection Network  & R & $8$ & $512$ & $26$ & $6$ & $5$ & \checkmark & \checkmark \\
$19$ & Optical Interconnection Network  & R & $5$ & $512$ & $26$ & $6$ & $5$ & \checkmark & \checkmark \\
$20$ & Credit Approval & C & $-1$ & $552$ & $31$ & $4$ & $8$ &  &  \\
$21$ & blood transfusion & C & $-1$ & $598$ & $4$ & $4$ & $0$ &  &  \\
$22$ & QSAR Bioconcentration classes dataset & R & $-1$ & $623$ & $29$ & $9$ & $5$ & \checkmark &  \\
$23$ & wiki4HE & R & $-10$ & $696$ & $284$ & $1$ & $50$ & \checkmark &  \\
$24$ & wiki4HE & R & $-11$ & $704$ & $284$ & $1$ & $50$ & \checkmark &  \\
$25$ & QSAR fish toxicity & R & $-1$ & $726$ & $18$ & $6$ & $2$ &  &  \\
$26$ & Tic-Tac-Toe Endgame & C & $-1$ & $766$ & $27$ & $0$ & $9$ &  &  \\
$27$ & QSAR biodegradation & C & $-1$ & $844$ & $123$ & $38$ & $15$ &  &  \\
$28$ & mirichoi0218\_insurance & R & $-1$ & $1070$ & $15$ & $3$ & $4$ &  &  \\
$29$ & Communities and Crime & R & $-1$ & $1595$ & $106$ & $99$ & $2$ &  &  \\
$30$ & Jasmine & C & $0$ & $2387$ & $144$ & $8$ & $136$ &  &  \\
$31$ & Abalone & R & $-1$ & $3341$ & $10$ & $7$ & $1$ & \checkmark &  \\
$32$ & mercedes-benz-greener-manufacturing & R & $1$ & $3367$ & $379$ & $0$ & $359$ &  &  \\
$33$ & Sylvine & C & $0$ & $4099$ & $20$ & $20$ & $0$ &  &  \\
$34$ & christine & C & $0$ & $4333$ & $1628$ & $1599$ & $14$ &  &  \\
$35$ & arashnic\_marketing-seris-customer-lifetime-value & R & $2$ & $6479$ & $72$ & $7$ & $15$ &  &  \\
$36$ & Seoul Bike Sharing Demand & R & $1$ & $7008$ & $15$ & $9$ & $3$ & \checkmark &  \\
$37$ & Electrical Grid Stability Simulated Data  & R & $-2$ & $8000$ & $13$ & $12$ & $1$ &  & \checkmark \\
$38$ & snooptosh\_bangalore-real-estate-price & R & $-1$ & $10656$ & $6$ & $2$ & $1$ &  &  \\
$39$ & MAGIC Gamma Telescope & C & $-1$ & $15216$ & $10$ & $10$ & $0$ &  &  \\
$40$ & Appliances energy prediction & R & $2$ & $15788$ & $25$ & $25$ & $0$ &  &  \\
$41$ & Nomao & C & $-1$ & $27572$ & $272$ & $116$ & $45$ &  &  \\
$42$ & Beijing PM2.5 Data & R & $5$ & $33405$ & $31$ & $10$ & $3$ &  &  \\
$43$ & Physicochemical Properties of Protein Tertiary Structure & R & $6$ & $36584$ & $9$ & $9$ & $0$ &  & \checkmark \\
\hline
\end{tabular}
\end{table}

\begin{table}[http!]
\label{tab:architectureinfos}
\centering
\tiny
\begin{tabular}{|l||c|c|c|c|c|}
\hline
id & name & task & target & source & file name \\
 &  &  & index &  &  \\
\hline
\hline
$0$ & Cervical Cancer Behavior Risk & C & $-1$ & archive.ics.uci.edu & sobar-72.csv \\
$1$ & Concrete Slump Test & R & $-1$ & archive.ics.uci.edu & slump\_test.data \\
$2$ & Concrete Slump Test & R & $-2$ & archive.ics.uci.edu & slump\_test.data \\
$3$ & Concrete Slump Test & R & $-3$ & archive.ics.uci.edu & slump\_test.data \\
$4$ & Breast Cancer Coimbra & C & $-1$ & archive.ics.uci.edu & dataR2.csv \\
$5$ & Algerian Forest Fires Dataset Sidi-Bel Abbes & C & $-1$ & archive.ics.uci.edu & Algerian\_forest\_fires\_dataset\_UPDATE.csv \\
$6$ & Algerian Forest Fires Dataset Bejaia & C & $-1$ & archive.ics.uci.edu & Algerian\_forest\_fires\_dataset\_UPDATE.csv \\
$7$ & restaurant-revenue-prediction & R & $-1$ & www.kaggle.com & train.csv.zip \\
$8$ & Servo & R & $-1$ & archive.ics.uci.edu & servo.data \\
$9$ & Computer Hardware & R & $-1$ & archive.ics.uci.edu & machine.data \\
$10$ & Breast Cancer & C & $0$ & archive.ics.uci.edu & breast-cancer.data \\
$11$ & Heart failure clinical records & C & $-1$ & archive.ics.uci.edu & heart\_failure\_clinical\_records\_dataset.csv \\
$12$ & Yacht Hydrodynamics & R & $-1$ & archive.ics.uci.edu & yacht\_hydrodynamics.data \\
$13$ & Ionosphere & C & $-1$ & archive.ics.uci.edu & ionosphere.data \\
$14$ & Congressional Voting Records & C & $0$ & archive.ics.uci.edu & house-votes-84.data \\
$15$ & Cylinder Bands & C & $-1$ & archive.ics.uci.edu & bands.data \\
$16$ & QSAR aquatic toxicity & R & $-1$ & archive.ics.uci.edu & qsar\_aquatic\_toxicity.csv \\
$17$ & Optical Interconnection Network  & R & $7$ & archive.ics.uci.edu & optical\_interconnection\_network.csv \\
$18$ & Optical Interconnection Network  & R & $8$ & archive.ics.uci.edu & optical\_interconnection\_network.csv \\
$19$ & Optical Interconnection Network  & R & $5$ & archive.ics.uci.edu & optical\_interconnection\_network.csv \\
$20$ & Credit Approval & C & $-1$ & archive.ics.uci.edu & crx.data \\
$21$ & blood transfusion & C & $-1$ & www.openml.org & php0iVrYT \\
$22$ & QSAR Bioconcentration classes dataset & R & $-1$ & archive.ics.uci.edu & Grisoni\_et\_al\_2016\_EnvInt88.csv \\
$23$ & wiki4HE & R & $-10$ & archive.ics.uci.edu & wiki4HE.csv \\
$24$ & wiki4HE & R & $-11$ & archive.ics.uci.edu & wiki4HE.csv \\
$25$ & QSAR fish toxicity & R & $-1$ & archive.ics.uci.edu & qsar\_fish\_toxicity.csv \\
$26$ & Tic-Tac-Toe Endgame & C & $-1$ & archive.ics.uci.edu & tic-tac-toe.data \\
$27$ & QSAR biodegradation & C & $-1$ & archive.ics.uci.edu & biodeg.csv \\
$28$ & mirichoi0218\_insurance & R & $-1$ & www.kaggle.com & insurance.csv \\
$29$ & Communities and Crime & R & $-1$ & archive.ics.uci.edu & communities.data \\
$30$ & Jasmine & C & $0$ & www.openml.org & file79b563a1a18.arff \\
$31$ & Abalone & R & $-1$ & archive.ics.uci.edu & abalone.data \\
$32$ & mercedes-benz-greener-manufacturing & R & $1$ & www.kaggle.com & train.csv.zip \\
$33$ & Sylvine & C & $0$ & www.openml.org & file7a97574fa9ae.arff \\
$34$ & christine & C & $0$ & www.openml.org & file764d5d063390.csv \\
$35$ & arashnic\_marketing-seris-customer-lifetime-value & R & $2$ & www.kaggle.com & squark\_automotive\_CLV\_training\_data.csv \\
$36$ & Seoul Bike Sharing Demand & R & $1$ & archive.ics.uci.edu & SeoulBikeData.csv \\
$37$ & Electrical Grid Stability Simulated Data  & R & $-2$ & archive.ics.uci.edu & Data\_for\_UCI\_named.csv \\
$38$ & snooptosh\_bangalore-real-estate-price & R & $-1$ & www.kaggle.com & blr\_real\_estate\_prices.csv \\
$39$ & MAGIC Gamma Telescope & C & $-1$ & archive.ics.uci.edu & magic04.data \\
$40$ & Appliances energy prediction & R & $2$ & archive.ics.uci.edu & energydata\_complete.csv \\
$41$ & Nomao & C & $-1$ & www.openml.org & phpDYCOet \\
$42$ & Beijing PM2.5 Data & R & $5$ & archive.ics.uci.edu & PRSA\_data\_2010.1.1-2014.12.31.csv \\
$43$ & Physicochemical Properties of Protein Tertiary Structure & R & $6$ & archive.ics.uci.edu & CASP.csv \\
\hline
\end{tabular}
\caption{Dataset description}
\end{table}

\subsubsection{Pre-processing.} 
\label{app:pre-processing}
To avoid biasing the benchmark towards specific methods and to get a result as general as possible, we only applied as little pre-processing as we could, without using any feature augmentation scheme. The goal is not to get the best possible performance on a given dataset but to compare the methods on equal ground.
We first removed uninformative features such as sample index. Categorical features with more than 12 modalities were discarded as learning embeddings is out of the scope of this benchmark. We also removed samples with missing target. 

\paragraph{Target treatment.}
The target is centered and standardized via the function $\fR(\cdot)$. We remove the observation  when the value is missing.






\bigskip

\paragraph{Features treatment.}
The imputation treatment is done during processing. For categorical  features, \NAN{} Data may be considered as a new class. For numerical features, we replace missing values by the mean. Set $n_j=\#\Set(X_j)$ the number of distinct values taken by the feature $X_j$, We proceed as follows :

\begin{itemize}
    \item[$\bullet$] When $n_j=1$, the feature $X_j$ is irrelevant, we remove it.
    \item[$\bullet$] When $n_j=2$ (including potentially \NAN{} class), we
    perform numerical encoding of binary categorical features.
    \item[$\bullet$]  Numerical features with less than $12$ distinct values are also treated as categorical features  ($2<n_j\leq 12$). We apply one-hot-encoding. 
    \item[$\bullet$] Finally, categorical features with $n_j> 12$ are removed.
   
\end{itemize}

\subsection{Methods} 
\label{app:datamodellist}

\paragraph{Usual methods.}
 We evaluate \trainmet{} against a large collection of popular methods (\XGB{} \cite{Breiman1997_arcing,Friedman2001,Friedman2002}, \XGBoost{}, \Catboost{} \cite{Prokhorenkova2018}, \cite{guestrin2016}, \lightgbm{} \cite{Ke2017}, \RF{} and \XRF{} \cite{breiman2001,barandiaran1998random}, \SVM{} and kernel-based \cite{Chihchung2011}, \MLP{} \cite{Hinton89connectionistlearning}, \Enet{} \cite{Zou05}, \Ridge{} \cite{Hoerl1970}, \LAS{} \cite{tibshirani1996}, \texttt{Logistic regression} \cite{cox1958}, \MARS{} \cite{Friedman1991}, \texttt{CART}, \texttt{XCART} \cite{Breiman1984,gey2005model,Klusowski2020sparse} ). 
 
\paragraph{\Fast\Catboost{} hyperparameter set} 

\Catboost is one of the dominant method on \TD, but also one of the slowest in terms of training time. We felt it was appropriate to evaluate a second version of \Catboost with a set of hyper-parameters designed to cut training time without decreasing performance too much. Following the official documentation guidelines, we picked the following hyperparameters for \Fast\Catboost:
\begin{itemize}
    \item "iterations":100,
    \item "subsample":0.1,
    \item "max\_bin":32,
    \item "bootstrap\_type",
    \item "task\_type":"GPU",
    \item "depth":3.
\end{itemize}

\paragraph{\FFNN{} architectures} 
 
See Table~\ref{tab:architectureinfos-app} for the different \FFNN{} architectures included in our benchmark.

\begin{table}[h]
	\centering
	\footnotesize
\caption{\FFNN{} architectures included in our benchmark. Legend for Block type: L=Linear; Res=Residual,GL=Gated Linear; GR=Gated Residual}
\label{tab:architectureinfos-app}
\begin{tabular}{|l||l|c|c|l|c|c|c|c|}
\hline
Name  &   \begin{tabular}{@{}c@{}}Block \\ Type\end{tabular}  & \# Blocks &  \begin{tabular}{@{}c@{}}Block \\ Depth\end{tabular} & Activation & Width & \begin{tabular}{@{}c@{}}Iter/ \\ epochs\end{tabular} & \begin{tabular}{@{}c@{}}Batch \\ Size\end{tabular}& max lr\\
\hline
\hline
\regularnetfast & \Linear & $2$ & $1$ & \ReLu & $256$ & $200$ & $\min(n,2048)$ &  $1e-2$ \\
\MLP & \Linear & $2$ & $1$ & \ReLu & $512$ & $500$ & $\min(n,2048)$ & $1e-2$ \\
\regularnetbatchstandard& \Linear & $2$ & 1 & \ReLu & $512$ & $20$ & $256$ &  $1e-2$ \\
\regularnetfastselu & \Linear & $2$ & $1$ & \ReLu & $256$ & $200$ & $\min(n,2048)$ &  $1e-2$ \\
\regularnetselu & \Linear & $2$ & $1$ & \SELU & $512$ & $500$ & $\min(n,2048)$ & $1e-2$ \\
\regularnetresblock & \ResBlock & $2$ & $2$ & \ReLu & $512$ & $500$ & $\min(n,2048)$ & $1e-2$ \\
\regularnetbatchresblock & \ResBlock & $2$ & $2$ & \ReLu & $512$ & $50$ & $256$ &  $1e-2$ \\
\regularnetglu & \GLU & $3$ & $1$ & \ReLu/\sigmoid& $512$ & $500$ & $\min(n,2048)$ &  $1e-2$ \\
\hline
\end{tabular}

\end{table}
\section{Hardware Details}
We ran all benchmark experiments on a cluster node with an Intel Xeon CPU (Gold 6230 20 cores @ 2.1 Ghz) and an Nvidia Tesla V100 GPU. All the other experiments were run using an Nvidia 2080Ti.

\section{Experiment results}

\subsection{Experiments on Hyper-Parameter Optimization}
 We timed on a subset of datasets and random seeds the use of HPO \cite{zimmer-tpami21a, feurer2020auto} which is both outside of the considered usecase and irrelevant to \trainmet{} since this method \textbf{does not introduce new hyper-parameters requiring tuning}.

\section{Supplementary material}

\end{document}


\maketitle

\begin{abstract}
This supplementary material contains the exhaustive description of our experiments, our complete results as well as the code for replicability purpose.
\end{abstract}

\subsection*{Replicability}

Our Python code is released as an open source package for replication:
\href{https://github.com/anonymousNeurIPS2021submission5254/SupplementaryMaterial}{github/anonymousNeurIPS2021submission5254/}.

\subsection*{Configuration machine}

We ran our experiments using several setups and GPU's: 
\begin{itemize}
    \item Google Cloud Plateform: NVIDIA Tesla P100,
    \item Google Colab : NVIDIA Tesla TESLA K80 and NVIDIA Tesla TESLA T4,
    \item Personal Computer : NVIDIA RTX 2080 Ti and NVIDIA RTX 2080 MaxQ.
\end{itemize}



\section{State of the Art}
We complete here the review of the existing literature on deep learning on tabular data.


An interesting line of research proposes to transpose the "leverage weak learners" idea underlying ensemble methods into neural networks. \cite{Olson2018moderna} proposes an interpretation of fitted \FFNN{} as ensembles of relatively weakly correlated, low-bias sub-networks. Thus this paper provides some insight on the generalization ability of overparametrized \FFNN{} on small datasets. Their experiments concerns binary classification on the UCI dataset but they did not attempt to outperform ensemble methods as it was not the goal of this work. 

The paper \cite{Arora2020meuf} carried out a study of Neural Tangent Kernel (\NTK) induced by infinitely wide neural networks on small classification tasks. \NTK{} slightly outperforms \RF{} implemented as in \cite{Delgado14a} on small UCI data sets ($n\leq 5000$). \NTK{} performs well on small size ($n \leq 640$) subsets of the CIFAR-10 benchmark but is inferior to ResNet-34 for larger size. However their architecture does not cover the regression task. Moreover, the super-quadratic running time of \NTK{} limits its use in large scale learning tasks.

\NetDNF{} \cite{katzir2021netdnf} is an end-to-end \DL{} model to handle tabular data. Its architecture is designed to emulate Boolean formulas in decision making. However, \XGBoost{}  outperforms \NetDNF{} in their experiments.

\cite{Klambauer2017} proposes Self-Normalized Neural Networks (\SNN) based on the \SELU{} activation function to train very deep feed-forward neural networks more efficiently. \SNN{} architecture is motivated as it makes SGD more stable. However \SNN{} requires careful tuning of hyperparameters and does not outperform \SVM{} or \RF{} on the UCI database.

\section{The \MLR-\FFNN{}}\label{secMLRloss}

\subsection{The \MLR{} loss}

Recall 
$$\bH=\bH(\bth,\bla,\bx)=\bA^{L-1}\left[(\bA^{L-1})^\top\bA^{L-1}+\bla \I\right]^{-1}(\bA^{L-1})^\top\in\R^{n\times n},$$
Where $\bA^{L-1}$ denotes the last hidden layer.
\begin{mydef}[\MLR{} regression loss]
\label{MLRlossBigsetting}
Set $\bH=\bH(\bth,\bla,\bx)$. We draw $i.i.d.$ random vectors $\xi$ and $\left(\xi_t\right)_{t=1}^T$ distributed as $\cN(0_n,\I_n)$. Let $\left(\perm^t(\bY)\right)^T_{t=1}$ be $T$ independently drawn permutations of $\bY$. We set $\overline{\bY} = mean(\bY)$ and define the \MLR{} loss as

\begin{align*}
\MLR(\bth,\bla) &= \RMSE\,\left(\bY+\tbH \xi\,;\,\bH\bY\right)\\
&\hspace{1cm}+\frac{1}{T}\sum_{t=1}^T\left|\RMSE\,( \bY\,;\, \overline{\bY}\1_n) -  \RMSE\,\left(\perm^t(\bY)+\tbH \,\xi_t\,;\,\bH\,\perm^t(\bY)\right)\right|.
\end{align*}
\end{mydef}

\paragraph{The benefit of close-form regularization.} 

The replacement of the output layer with Ridge regularizes the network in two ways:
$(i)$ the weights on the output layer are a direct function of the last hidden layer $\bA^{L-1}$. This effect is much stronger than adding a constraint or a penalty directly on $W^L$ the weights of the $L$-th layer of the \FFNN; $(ii)$ the close-form we choose is the ridge instead of the OLS, which implicitly subjects the weights to a steerable $L_2$ regularization.

\paragraph{The generalization effect of random permutations.} Our work is loosely related to \cite{zhang2016understanding} where label permutations are used after the model has been trained as a \textit{qualitative} observational method to exhibit the overfitting capacity of Neural networks. In our approach, we go further as we use random permutations during the training phase to define a \textit{quantitative} measure of the amount of overfitting of a model. This measure is actively used to penalize overfitting during the training phase. This is the underlying mecanism behind the \MLR{} loss.
First, when we take a permuted label vector we obtain a new label vector with two properties. First both $\bY$ and $\perm(\bY)$ admit the same marginal distributions. This new vector can be seen as a "realistic" data-augmented new sample for the training set. Second the expected number of fixed points ($\perm(i) = i$) in a permutation drawn uniformly at random is equal to $1$ (See Chapter 5 in \cite{permutebook}); $i.e.$ the proportion of fixed points in a random permutation of $n$ elements is insignificant. Thus the label permutation breaks the dependence relationship between $Y_{\perm(i)}$ and $\bx_i$. Therefore, $\bx_i$ provides no information on the possible value of $Y_{\perm(i)}$ and predicting $Y_{\perm(i)}$ using $\bx_i$ can only result in overfitting. In other words, label permutation is used to produce a control set $(\bx,\perm(\bY))$ that can only be fitted through memorization. \MLR{} focuses on patterns that appear only in $(\bx,\bY)$ and not in uncorrelated pairs $(\bx,\perm(\bY))$.

\paragraph{Structured Dithering.} 
We describe an additional scheme to prevent memorization. We apply a dithering scheme which adapts to the spectral structure of $\bH$, the "regularized projector" based on $\bA^{L-1}$ (the output of the last hidden layer). More specifically, we muddle the target using $\tbH \xi\,$ which introduces noise of higher variance along the weakly informative eigendirections of $\bH$.

\subsection{Cross-Entropy loss}

In the classification task, the \FFNN{} architecture is essentially unchanged. The usual loss for binary classification task is the \BCE{} loss that combines a \sigmoid{} and the Cross Entropy (\CE) loss.
(namely \texttt{torch.nn.BCEWithLogitsLoss} in PyTorch and referred to as \BCE{} in this paper). 
Set $\s(\cdot)=\sigmoid(\cdot)$, then
$$\BCE(\bY, f(\bth,\bx))=-\frac1n\left[\bY^\top\log(\s(f(\bth,\bx))
+(\1_n-\bY)^\top\log(\1_n-\s(f(\bth,\bx))\right].$$

\bigskip

\begin{mydef}[\CEMLR{} loss]
\label{CElossBigsetting}
Let $\xi$ and $\left(\xi_t\right)_{t=1}^T$ be $i.i.d.$ $\cN(0_n,\I)$ vectors. Set $\bY^* = 2 \bY - 1$. We define the \CEMLR{} loss as
\begin{align*}
\CEMLR(\bth,\bla) &= \BCE\,\left(\bY;\,\bY^*+\tbH \xi+\bH\bY^*\right)
\\
&\hspace{0.25cm}+\frac{1}{T}\sum_{t=1}^T\left|\BCE\,( \bY\,;\, \overline{\bY}\1_n) -\BCE\,\left(\perm^t(\bY^*) \,;\,  \perm^t(\bY^*)+\tbH \xi_t+\bH\,\perm^t(\bY^*)\right)\right|.
\end{align*}
\end{mydef}

\bigskip

The quantity $\BCE\,( \bY\,;\, \overline{\bY}\1_n)$ is our \baseline. Note that $\bY^*$ with values in $\{-1,1\}$ is the symmetrized version of $\bY$. Next, the Structured dithering is applied to the prediction rather than the target $\bY$ because the \BCE{} is only defined for binary target $\bY \in \{0;1\}^n$. 

\begin{mydef}[\CENN]
Our \CEMLR{} neural net (\CENN) is 
\begin{align*}
    \label{CENN}
 &\CENN(\hbth,\hatla,\bigcdot)=\Hd(\bA^{L-1}(\hbth,\bigcdot) \,\bW(\hbth,\hatla,\bx)) \bY \in\{0,1\}^{obs.}\noindent\\
&\text{with}\,\, (\hbth,\hatla) = \underset{\bth,  \bla}{\arg\min} \;\CEMLR(\bth,\bla),
\end{align*}
and $\bW(\cdot,\cdot,\bx)$ $s.t.$
, $\bW=\bW(\bth,\bla,\bx)=\left[(\bA^{L-1})^\top\bA^{L-1}+\bla \I\right]^{-1}(\bA^{L-1})^\top\in\R^{J\times n}$. 
\end{mydef}

\bigskip

\section{Training a \FFNN{} with \MLR}

\paragraph{The \NN{} Architecture.} 
We consider \FFNN{} with $L$ layers, $L\in\cG_L:=\{1,2,3,4\}$, and with all the hidden layers of constant width $J$. In our experiments, we always take $J$ as large as possible (our machine with 11GVRAM allowed for $J=2^{10}$).

\begin{table}[H]
	\caption{Benchmarked architectures. }
	\label{tab:architectures1}
\footnotesize
 	\centering
 \begin{tabular}{|c||c|c |c|c|c |c|c|c|c|}
 \hline
Architecture & $L$ & $\lr$ & $\max_{\iter}$ & \fixB & $J$ & $\bs$ & $T$ & \multicolumn{2}{|c|} {$\widetilde\sigma$ }   \\ 
\hline
 $\MLRa $& $1$ & $10^{-2}$ & $200$ & $10'$ & $2^{10}$ & $min(n, J)$ & $16$ & Reg.: $0.03$ & Classif.: $0$ \\
  \cline{9-10} 
 $\MLRb $& $2$ & $10^{-3}$ & $200$ & $id.$ & $id.$ & $id.$& $id.$ &\multicolumn{2}{|c|} {$id.$}\\
 $\MLRc $& $3$ & $10^{-3.5}$ & $400$ &  $id.$ & $id.$ & $id.$& $id.$ & \multicolumn{2}{|c|} {$id.$}\\
 $\MLRd $& $4$ & $10^{-4}$ & $400$ &  $id.$ & $id.$ & $id.$& $id.$ & \multicolumn{2}{|c|} {$id.$} \\
 \hline
 \end{tabular}
 \end{table}
 

\paragraph{\Dither{} \cite{dither}.} 
This step is distinct from the Structured dithering that we introduced in the \MLR{} method.
In the regression setting, we do not apply the \MLR{} loss on $\bY$ but rather on a noisy version of $\bY$ as is usually done in practice. Let $\epsilon,\,(\epsilon^t)_t\,\overset{i.i.d}{\sim}\cN(\bO,\tilde\sigma^2\I)$. We set $\bY_{\eps}=\bY+\eps$ and
$\perm_{\epsilon}^t(\bY)=\perm^t(\bY)+\epsilon^{t}$.
In our experiments, we use the \MLR{} loss on $\left(\bY_{\epsilon},\left(\perm_{\epsilon}^t(\bY)\right)_{t=1}^T\right)$ instead of $\left(\bY,\left(\perm^t(\bY)\right)_{t=1}^T\right)$.
\begin{align*}
\MLR(\bth,\bla) &= \RMSE\,\left(\bY_{\eps}+\tbH \xi\,;\,\bH\bY_{\eps}\right)\\
&\hspace{1cm}+\frac{1}{T}\sum_{t=1}^T\left|\RMSE\,( \bY\,;\, \overline{\bY}\1_n) -  \RMSE\,\left(\perm_{\eps}^t(\bY)+\tbH \,\xi_t\,;\,\bH\,\perm_{\eps}^t(\bY)\right)\right|.
\end{align*}

Here again, $\tilde\sigma$ is \textbf{not an hyperparameter} as we use the same value $\tilde\sigma=0.03$ for all the datasets in our benchmark.  Moreover, in our approach the batch size $\bs$ is not a hyperparameter as we fix it as in table above.

Note that we do not apply this dither step in the classification setting.

\paragraph{Initialization of $\bth$.}
 The initialization of $\bth$ is as in \cite{Gloriotetal}.

\medskip

\begin{mdframed}
\underline{Recall $|\Input|=d$ and $|\out|$=1}
\medskip

$\forall\el\in\llbracket1,L-1 \rrbracket$, $b^{\el}=\bO$. The entries of $W^\el$ are generated independently from the uniform distribution on the interval $\cI_\ell$ :  
\begin{itemize}
   \item[$\bullet$] $\cI_1=\left(-\sqrt{\frac{6}{(d+J)}},\sqrt{\frac{6}{(d+J)}}\right)$ and $\,\cI_L=\left(-\sqrt{\frac{6}{(d+1)}},\sqrt{\frac{6}{(d+1)}}\right)$
   \item[$\bullet$] $\cI_\el=\left(-\sqrt{\frac{6}{(J+J)}},\sqrt{\frac{6}{(J+J)}}\right)$, $\forall\el\in\llbracket2,L-1 \rrbracket $
\end{itemize}
\end{mdframed}

\bigskip

\paragraph{Efficient heuristic to initialize the Ridge parameter.} 
In our experiments, we pick $\bla_{init}$ by running a grid-search on the finite difference approximation for the derivative of \MLR{} on the grid 
$\cG_{\bla}=\left\lbrace \bla^{(k)} = 10^{-1}\times10^{5 \times k / 11}\;: \;   k = 0, \cdots, 11 \right\rbrace$:
\begin{eqnarray*}
\bla_{init}=\sqrt{\bla^{(\hat {k}})\,\bla^{(\hat {k}+1)}}\;\text{where}\;\hat {k}=\arg\max\left\{ \left(\MLR(\bth,\bla^{(k+1)}) - \MLR(\bth,\bla^{{(k)}})\right),\, \bla^{(k)}\in\cG_{\bla}\right\}.
\end{eqnarray*}
The Ridge parameter $\bla$ is \textbf{not an hyperparameter} of our method; it is trained alongside the weights of the Neural Net architecture.

\bigskip
\paragraph{Choice of the number of iterations during the train.}
\begin{itemize}
    \item [$\bullet$] We fix the maximum number of iterations $\texttt{max}_{\iter}$ (depending on the value of $L$).
    \item [$\bullet$]  We fix the budget (\fixB = 5 min) and denote by $n_{\iter}$ the possible number of iterations during the allotted time \fixB. 
    \item [$\bullet$] We denote by $\iter$ the number of iterations that will actually be performed, $i.e.$
$$\iter= \min(\texttt{max}_{\iter},n_{\iter})$$
\end{itemize}

 \bigskip
  
\paragraph{Training \MLR{}-NN.}

We train the \FFNN{} with $\bs=\min(J,n)$ and we use Adam \cite{kingma2014adam} with default parameters except for the learning rate $\lr$ which depends on the number of layers $L$ (Table~\ref{tab:architectures1}). 
\medskip

\begin{mdframed}

$
\begin{array}{l}
    \textbf{Training}\\
      \quad \left|\begin{array}{llll}
          \textbf{Initialization}\\
                \quad \left| \begin{array}{ll}
                   \Set\,\bth \\
                     \Set\,\bla \\
                \end{array} \right.\\
    \textbf{Optimization}\\
    \quad \left| \begin{array}{ll}
         \While\,\, e < \iter \,\,\,\Do:\\
         \quad \left| \begin{array}{llllll}
           \bA^{0}-> \bx\in\R^{\bs\times d}\\
          \For\,\, \ell = 1 \cdots L-1:\\
               \quad \left|\begin{array}{l}
                           \bA^{\el} ->
                     \ReLu(\bA^{\el-1}W^{\el} +B^{\el})
                       \end{array} \right.\\
             \bH(\bth,\bla) -> \bA^{L-1}\left[(\bA^{L-1})^\top\bA^{L-1}+\bla \I_J\right]^{-1}{\bA^{L-1}}^\top\\
             \texttt{Compute }\, \MLR(\bth,\bla) \quad \text{or}\quad \CEMLR(\bth,\bla)
             \\
            \textbf{Backpropagate }\, (\bth,\bla) \textbf{ through }\, \MLR(\bth,\bla) \quad \text{or}\quad \CEMLR(\bth,\bla)\\
            e -> e+1\\
        \end{array} \right.\\
    \end{array} \right.\\
    \end{array} \right.\\
\end{array}
$
\end{mdframed}

We select a $validation$-set of size $n_{val}=20\%\, n$. We read the $\Rdeux$-score for each iteration on the $validation$-set and take the iteration with the best $\Rdeux$-score: 
$$\iter^*:=\arg\max\{\texttt{R}^2_k, \ k =1,\cdots,\iter\}.$$
Finally, $(\hbth,\hatla)$ will take its value at iteration $\iter^*$

\bigskip

\paragraph{Our final models.}
We propose several models with varying depth based on \FFNN{} trained with the \MLR{} loss. We also create ensemble models combining architectures of different depth.
Our models are:\\
$\bullet$  \MLRL: a simple \FFNN{} of depth $L$ ($1\leq L\leq 4$).\\
$\bullet$ \BMLRL: a bagging of 10 \FFNN{} of depth $L$ ($L=1$ or $L=2$).\\
$\bullet$ \BMLR{}: an ensemble of 20 \FFNN{} (the aggregation of \BMLRa{} and \BMLRb{} of depth $1$ and $2$ respectively).\\
$\bullet$ \BestMLR{}: the best prediction among 20 \NN{} in terms of the validation score.\\
$\bullet$ \TopfiveMLR{}: the aggregation of the top 5 among 20 \NN{} in terms of the validation score.

For the methods based on bagging \cite{breiman1996}, the final prediction is the mean of each \NN{} prediction.

\section{Construction of the Benchmark}


To produce this benchmark (Table~\ref{tab:datasets}), we aggregated 32 tabular datasets (16 in regression  and 16 in classification), from the UCI repository and Kaggle. 
For computational reasons, we have chosen to restrict the number of datasets but we performed more $train$/$test$ splitting in order to reduce the variance of our results. 
We curated the UCI repository and Kaggle through a set of rules ($e.g.$ discard empty or duplicate datasets, times series, missing target, non $i.i.d.$ samples, text format, $etc$.).
\begin{table}[H]
\caption{Benchmark datasets. \# Num. and \# Cat. denote the initial number of numerical and categorical features respectively. We denote by $d$ the number of features after the pre-processing and one-hot encoding.}
\label{tab:datasets}
\centering
\footnotesize
\begin{tabular}{|l|c|c|c|c|c|}
\hline
Description & Task & $n$ & $d$ & \# Num. &  \# Cat.  \\
\hline
\hline
Concrete Slump Test -2 & \reg  & $ 103$ & $8$ & $8$ & $0 $ \\
\hline
Concrete Slump Test -3 & \reg  & $ 103$ & $8$ & $8$ & $0 $ \\
\hline
Concrete Slump Test -1 & \reg  & $ 103$ & $8$ & $8$ & $0 $ \\
\hline
Servo & \reg  & $ 168$ & $24$ & $2$ & $4 $ \\
\hline
Computer Hardware & \reg  & $ 210$ & $7$ & $7$ & $0 $ \\
\hline
Yacht Hydrodynamics & \reg  & $ 308$ & $33$ & $5$ & $3 $ \\
\hline
QSAR aquatic toxicity & \reg  & $ 546$ & $34$ & $8$ & $3 $ \\
\hline
QSAR Bioconcentration classes  & \reg  & $ 779$ & $25$ & $8$ & $4 $ \\
\hline
QSAR fish toxicity & \reg  & $ 909$ & $18$ & $6$ & $2 $ \\
\hline
insurance & \reg  & $ 1338$ & $15$ & $3$ & $4 $ \\
\hline
Communities and Crime & \reg  & $ 1994$ & $108$ & $99$ & $2 $ \\
\hline
Abalone R & \reg  & $ 4178$ & $11$ & $7$ & $1 $ \\
\hline
squark automotive CLV training & \reg  & $ 8099$ & $77$ & $7$ & $16 $ \\
\hline
Seoul Bike Sharing Demand & \reg  & $ 8760$ & $15$ & $9$ & $3 $ \\
\hline
Electrical Grid Stability Simu & \reg  & $ 10000$ & $12$ & $12$ & $0 $ \\
\hline
blr real estate prices & \reg  & $ 13320$ & $2$ & $2$ & $0 $ \\
\hline
Cervical Cancer Behavior Risk & \Clf  & $ 72$ & $149$ & $19$ & $14 $ \\
\hline
Post-Operative Patient & \Clf  & $ 91$ & $32$ & $0$ & $8 $ \\
\hline
Breast Cancer Coimbra & \Clf  & $ 116$ & $9$ & $9$ & $0 $ \\
\hline
Heart failure clinical records & \Clf  & $ 299$ & $12$ & $7$ & $5 $ \\
\hline
Ionosphere & \Clf  & $ 352$ & $34$ & $32$ & $2 $ \\
\hline
Congressional Voting Records & \Clf  & $ 436$ & $64$ & $0$ & $16 $ \\
\hline
Cylinder Bands & \Clf  & $ 541$ & $111$ & $1$ & $19 $ \\
\hline
Credit Approval & \Clf  & $ 691$ & $42$ & $4$ & $8 $ \\
\hline
Tic-Tac-Toe Endgame & \Clf  & $ 959$ & $36$ & $0$ & $9 $ \\
\hline
QSAR biodegradation & \Clf  & $ 1056$ & $141$ & $41$ & $15 $ \\
\hline
Chess (King-Rook vs. King-Pawn & \Clf  & $ 3196$ & $102$ & $0$ & $36 $ \\
\hline
Mushroom & \Clf  & $ 8125$ & $125$ & $0$ & $21 $ \\
\hline
Electrical Grid Stability Simu & \Clf  & $ 10000$ & $12$ & $12$ & $0 $ \\
\hline
MAGIC Gamma Telescope & \Clf  & $ 19021$ & $10$ & $10$ & $0 $ \\
\hline
Adult & \Clf  & $ 32561$ & $34$ & $6$ & $5 $ \\
\hline
Internet Firewall Data & \Clf  & $ 65532$ & $11$ & $11$ & $0 $ \\
\hline
\end{tabular}
\end{table}

\subsection{Pre-processing}

To avoid biasing the benchmark towards specific methods and to get a result as general as possible, we only applied as little preprocessing as we could, without using any feature augmentation scheme. The goal is not to get the best possible performance on a given dataset but to compare the methods on equal ground.
We first removed uninformative features such as sample index. Categorical features with more than 12 modalities were discarded as learning embeddings is out of the scope of this benchmark. We also removed samples with missing target.

\paragraph{Target treatment.}
The target is centered and standardized via the function $\fR(\cdot)$. We remove the observation  when the value is missing.

\begin{mdframed}
$
\begin{array}{l}
\fR(Y)\\
\quad \left|\begin{array}{ll}
Y -> \text{float32}(Y)\\
\textbf{for}\,\,  i=1:n\\
\quad \left| \begin{array}{l}
\If\,\, Y_i==\NAN \\
\qquad\Remove(x_i,Y_i)\\
\end{array} \right.\\
Y -> \frac{Y-\overline{Y}}{\bar\sigma(Y)}\\
\end{array} \right.
\end{array}
$
\end{mdframed}

\bigskip

\paragraph{Features treatment.}
The imputation treatment is done during processing. For categorical  features, \NAN{} Data may be considered as a new class. For numerical features, we replace missing values by the mean. Set $n_j=\#\Set(X_j)$ the number of distinct values taken by the feature $X_j$, We proceed as follows :

\begin{itemize}
    \item[$\bullet$] When $n_j=1$, the feature $X_j$ is irrelevant, we remove it.
    \item[$\bullet$] When $n_j=2$ (including potentially \NAN{} class), we
    perform numerical encoding of binary categorical features.
    \item[$\bullet$]  Numerical features with less than $12$ distinct values are also treated as categorical features  ($2<n_j\leq 12$). We apply one-hot-encoding. 
    \item[$\bullet$] Finally, categorical features with $n_j> 12$ are removed.
\end{itemize}

\subsection{Compared methods}
\label{sec: ComparedMethod}


We ran the benchmark with \textbf{all the methods} (see Table~\ref{tab:methods}) \textbf{available in the scikit-learn library} for classification and regression (including \RF{} and \XGB) as well as the \GBDT{} methods. All methods were ran with the default hyperparameters. 

\begin{table}[H]
\caption{Main classes of methods.}
\label{tab:methods}
	\centering
	\footnotesize
\begin{tabular}{|l|l|}
\hline
Class      & \\
of Methods & Methods\\
\hline
\MLR{} (this paper)   & \MLRL, \BMLRL, \BMLR, \BestMLR, \TopfiveMLR  \\
\hline
\GBDT  & \XGB{} \cite{Breiman1997_arcing,Friedman2001,Friedman2002},  \Catboost{} \cite{Prokhorenkova2018}, \XGBoost{} \cite{guestrin2016}, \lightgbm{} \cite{Ke2017} \\
\hline
\RF  & \RF{} and \XRF{} \cite{breiman2001,barandiaran1998random}  \\
\hline               
\SVM &  \texttt{Lin-}\SVM{}, \SVM{}, $\nu$\texttt{-}\SVM{} \cite{Chihchung2011}\\
\hline
\RNN   & \texttt{Fast.ai} \cite{Howard2020}, \MLP \cite{Hinton89connectionistlearning}\\
\hline
\GLM   & \texttt{OLS}, \Enet{} \cite{Zou05}, \Ridge{} \cite{Hoerl1970}, \LAS{} \cite{tibshirani1996}, \texttt{Logistic regression} \cite{cox1958}\\
\hline
\MARS & \MARS{} \cite{Friedman1991}\\
\hline
\TREE& \texttt{CART}, \texttt{XCART} \cite{Breiman1984,gey2005model,Klusowski2020sparse}\\
\hline
\Baseline & Reg: \texttt{Intercept}$\,|\,$ Classif: \texttt{Class probabilities}\\
\hline
\end{tabular}
\end{table}

\section{\MLR{} Parameters Analysis} 

In this section we study the behavior of the \MLR{} method and the impact of its key components through extensive evaluation on three datasets, \UCIa, \UCIb{} and \UCIc{}, for which $(n,d)$ are equal to $(103,8)$, $(546,34)$ and $(8760,15)$ respectively. We repeated each experiment over 100 random $train$/$test$ splits.

\subsection{Impact of the MLR components.}

In this section, we study the impact of the different components in the \MLR{} approach on the
the $\Rdeux$-score on the $test$ and $validation$ sets, computation time, the  convergence of the method (\Iter) and the initialization of the Ridge parameter $\bla_{init}$. To study the impact of each specific parameter, we set the other ones equal to their default values in Table~\ref{tab:architectures1}. Note that for the following study, we chose a batch size $\bs=\min(n,2^{14})$, unlike in our main experiments where we took $\bs=\min(n,2^{10})$ due to time constraints.

Note also that due to access failure to \VM, computation time was sometimes obtained on a less powerful configuration in Tables \ref{tab:StructDithering}, \ref{tab:Permutation}, \ref{tab:lambdaInit} and \ref{tab:Dithering}. We marked by an asterisk $\boldsymbol{\ast}$ any computation time obtained on the \Portable{} configuration.

\paragraph{Structured Dithering.} 
Recall that we added Structured noise $(\I_n-\bH )\xi$ to the target $\bY$ with $\xi\sim \cN(0,\sigma^2\I)$.
Table~\ref{tab:StructDithering} reveals the impact of the structured dithering parameter $\sigma$. Default value ($\sigma = 1$) yields consistently good generalization performance. Of course, it is always possible to tune this hyperparameter around value $1$ for potential improvement of the generalization performances. Higher values of $\sigma$ lead to a significant degradation of $\Rdeux$-score as it caused the method to diverge. In our experiments, $\sigma$ was not an hyperparameter as it was always set equal to $1$. Moreover, adding structured dithering has no impact on the value of $\bla_{init}$ or computational time.

\begin{table}[H]
\caption{Structured dithering dependence.}
\label{tab:StructDithering}
	\centering
\begin{tabular}{|c|c|c|c|c|c|c|}
\hline
 \UCIa &$\sigma$ &  $\Rdeux$ & Time & \iter & $\Rdeux_{val}$ & $\bla_{init}$  \\
\hline
\hline
&0  & $ 0.321$ & $30.356$ & $60.650$ & $0.479$ & $219.345$ \\
\hline
&0.2  & $ 0.338$ & $30.424$ & $79.830$ & $0.496$ & $219.345 $ \\
\hline
&\textbf{1} & $ 0.357$ & $30.423$ & $99.570$ & $0.515$ & $219.345 $ \\
\hline
&2 & $ 0.089$ & $1.312$ & $0.250$ & $0.137$ & $219.345 $ \\
\hline
&3 & $ 0.068$ & $1.257$ & $0.000$ & $0.116$ & $219.345 $ \\
\hline
\hline
\UCIb & $\sigma$ &  $\Rdeux$  & Time & \iter & $\Rdeux_{val}$  & $\bla_{init}$  \\
\hline
\hline
&0  & $ 0.463$ & $32.250$ & $11.200$ & $0.511$ & $774.264 $ \\
\hline
&0.2  & $ 0.463$ & $32.408$ & $14.550$ & $0.514$ & $774.264 $ \\
\hline
&\textbf{1} & $ 0.460$ & $32.281$ & $46.750$ & $0.525$ & $774.264 $ \\
\hline
&2 & $ 0.220$ & $1.276$ & $0.020$ & $0.226$ & $774.264 $ \\
\hline
&3 & $ 0.216$ & $1.288$ & $0.000$ & $0.223$ & $774.264 $ \\
\hline
\hline
 \UCIc &$\sigma$  &  $\Rdeux$  & Time & \iter & $\Rdeux_{val}$  & $\bla_{init}$  \\
\hline
\hline
&0  & $ 0.863$ & $89.425$ & $181.300$ & $0.864$ & $10000.001 $ \\
\hline
&0.2  & $ 0.863$ & $90.206$ & $188.520$ & $0.864$ & $10000.001 $ \\
\hline
&\textbf{1} & $ 0.855$ & $89.968$ & $191.920$ & $0.857$ & $10000.001 $ \\
\hline
&2  & $ 0.364$ & $1.876$ & $0.000$ & $0.363$ & $10000.001 $ \\
\hline
&3 & $ 0.364$ & $1.891$ & $0.000$ & $0.363$ & $10000.001 $ \\
\hline
\end{tabular}
\end{table}

\paragraph{Permutations.}
We studied the impact of the randomness aspect of the \MLR{} loss. We compared different sets of permutations drawn at random. The choice of the seed has little impact on the value of the \MLR{} loss as soon as $T\geq 2^2$. Table~\ref{tab:Permutation} reveals a significant jump in $\Rdeux$-score on the test going from $T=0$ to $T=1$ permutation. Then, increasing the value of $T>1$ may sometimes slightly improve $\Rdeux$-score.
Meanwhile, a larger number of permutations has a direct negative impact on runtime per iteration and VRAM footprint. Past a certain threshold $2^8$, GPU parallelization no longer prevents the linear dependency on $T$. We escape any trade-off by picking $T=2^4$ permutations in all our experiments. This value is large enough for the \MLR{} loss to converge (with regards to $T$), yet still leveraging GPU parallelization. 

\begin{table}[H]
	\caption{Permutation dependence. $\boldsymbol{\ast}$: computation time was obtained with a \Portable.}
		\label{tab:Permutation}
	\centering
\begin{tabular}{|c|c|c|c|c|c|c|}
\hline
\UCIa & $T$  &  $\Rdeux$ & Time & \iter & $\Rdeux_{val}$ & $\bla_{init}$  \\
\hline
\hline
 &  0  & $ 0.252$ & $3.184$ & $61.050$ & $0.401$ & $31.831 $ \\
\hline
 &  1  & $ 0.331$ & $3.357$ & $110.040$ & $0.459$ & $285.238 $ \\
\hline
 &  2  & $ 0.338$ & $3.359$ & $109.960$ & $0.468$ & $215.370 $ \\
\hline
 &  $2^{2}$  & $ 0.343$ & $3.358$ & $109.370$ & $0.473$ & $219.345 $ \\
\hline
 &  $\boldsymbol{2^{4}}$  & $ 0.347$ & $4.012^{\textbf{*}}$ & $116.190$ & $0.484$ & $216.235 $ \\
\hline
 &  $2^{8}$  & $ 0.351$ & $3.371$ & $117.160$ & $0.494$ & $219.345 $ \\
\hline
 &  $2^{10}$  & $ 0.349$ & $3.433$ & $117.650$ & $0.495$ & $219.345 $ \\
\hline
\hline
\UCIb & $T$  &  $\Rdeux$ & Time & \iter & $\Rdeux_{val}$ & $\bla_{init}$  \\
\hline
\hline
 &  0.0  & $ 0.460$ & $3.253$ & $46.770$ & $0.509$ & $774.264 $ \\
\hline
 &  1  & $ 0.461$ & $3.452$ & $62.020$ & $0.518$ & $774.264 $ \\
\hline
 &  2  & $ 0.466$ & $3.461$ & $60.040$ & $0.518$ & $774.264 $ \\
\hline
 &  $2^{2}$  & $ 0.469$ & $3.462$ & $60.720$ & $0.521$ & $774.264 $ \\
\hline
 &  $\boldsymbol{2^{4}}$  & $ 0.473$ & $6.172^{\textbf{*}}$ & $72.800$ & $0.527$ & $774.264 $ \\
\hline
 &  $2^{8}$  & $ 0.477$ & $3.496$ & $81.900$ & $0.532$ & $774.264 $ \\
\hline
 &  $2^{10}$  & $ 0.480$ & $3.551$ & $81.530$ & $0.532$ & $774.264 $ \\
\hline
\hline
\UCIc & $T$ &  $\Rdeux$ & Time & \iter & $\Rdeux_{val}$ & $\bla_{init}$  \\
\hline
\hline
 &  0  & $ 0.817$ & $8.251$ & $197.830$ & $0.817$ & $10000.001 $ \\
\hline
 &  1  & $ 0.813$ & $8.606$ & $197.860$ & $0.813$ & $10000.001 $ \\
\hline
 &  2  & $ 0.813$ & $8.654$ & $197.400$ & $0.814$ & $10000.001 $ \\
\hline
 &  $2^{2}$  & $ 0.813$ & $8.645$ & $197.780$ & $0.814$ & $10000.001 $ \\
\hline
 &  $\boldsymbol{2^{4}}$  & $ 0.814$ & $30.654^{\textbf{*}}$ & $197.100$ & $0.814$ & $10000.001 $ \\
\hline
 &  $2^{8}$  & $ 0.813$ & $10.391$ & $197.230$ & $0.814$ & $10000.001 $ \\
\hline
 &  $2^{10}$  & $ 0.814$ & $17.330$ & $197.070$ & $0.814$ & $10000.001 $ \\
\hline
\end{tabular}
\end{table}

\paragraph{Initialization of Ridge parameter $\bla_{init}$.}
Recall that Ridge regularization is the essential component of the \MLR{} method as it provides a closed form representation of the last hidden layer on which we can conveniently apply the follow-up steps: structured dithering and random permutations. Contrary to $T$ and the dither parameter $\sigma$, the choice of the appropriate initial value of $\bla$ is very impactful and depends on both network architecture and dataset characteristics as shown in Table~\ref{tab:lambdaInit}.

When we compare the value $\bla_{init}$ given by our heuristic (in bold) with the other values chosen in Table~\ref{tab:lambdaInit}, we observe that our heuristic is quite effective, as in average on the 3 datasets, it is always within $3\%$ of the best value in the grid of Table \ref{tab:lambdaInit} in term of $\Rdeux$-score on the $test$. As we can see for the \UCIb{} dataset, the optimal value was not within the bounds of the grid $\cG_{\bla}$ we chose. Using a larger grid with a bigger granularity would improve the results.



Despite access failure to \VM{} for one specific value of $\bla_{init}$, our main experiments reveal a small runtime overcost for the initialization step, 
mostly because all steps including the matrix inversion need to be performed only once and do not require computing the derivation graph. We favored a small simple grid 
$\cG_{\bla}=\left\lbrace \bla^{(k)} = 10^{-1}\times10^{5 \times k / 11}\;: \;   k = 0, \cdots, 11 \right\rbrace$ to select $\bla_{init}$. This grid was designed to work well on small size datasets. Of course, it is possible to refine this grid with respect to the dataset size and architecture at hand to achieved even higher generalization performance.
Another possible approach could be to tune $\bla_{init}$ on the $validation$ set. Indeed, we observe in Table \ref{tab:lambdaInit} that the optimal value of $\bla_{init}$ on the $test$ seems to be close to that obtained on the $validation$ set.

\begin{table}[H]
	\caption{Dependence on $\bla_{init}$. $\boldsymbol{\ast}$: computation time was obtained with a \Portable.}
		\label{tab:lambdaInit}
	\centering
\begin{tabular}{|c|c|c|c|c|c|}
\hline
\UCIa &$\bla_{init}$ & $\Rdeux$ & Time & \iter & $\Rdeux_{val}$  \\
\hline
\hline
 &  0  & $ -0.110$ & $0.180$ & $7.790$ & $-0.020 $ \\
\hline
&$10^{-3}$  & $ -0.444$ & $2.078$ & $90.270$ & $0.265 $ \\
\hline
&$10^{-1}$  & $ 0.097$ & $2.083$ & $70.310$ & $0.254 $ \\
\hline
&$10$  & $ 0.320$ & $2.070$ & $116.630$ & $0.466 $ \\
\hline
 &  $\boldsymbol{216.235}$  & $ 0.347$ & $2.902^{\textbf{*}}$ & $116.190$ & $0.484 $ \\
\hline
&$10^{3}$  & $ 0.359$ & $2.087$ & $125.020$ & $0.480 $ \\
\hline
&$10^{5}$  & $ 0.334$ & $2.103$ & $152.460$ & $0.428 $ \\
\hline
&$10^{7}$  & $ 0.263$ & $2.104$ & $188.630$ & $0.339 $ \\
\hline
&$10^{9}$  & $ -0.050$ & $2.089$ & $197.890$ & $-0.009 $ \\
\hline
\hline
 \UCIb  &$\bla_{init}$ & $\Rdeux$ & Time & \iter & $\Rdeux_{val}$  \\
\hline
\hline
 &  0  & $ -0.276$ & $0.014$ & $0.010$ & $-0.244 $ \\
\hline
&$10^{-3}$  & $ -33.053$ & $0.133$ & $2.510$ & $-9.371 $ \\
\hline
&$10^{-1}$  & $ -3.768$ & $2.137$ & $36.770$ & $-0.151 $ \\
\hline
&$10$  & $ 0.422$ & $2.086$ & $9.530$ & $0.477 $ \\
\hline
 &  $\boldsymbol{774.263}$  & $ 0.473$ & $3.426^{\textbf{*}}$ & $72.800$ & $0.527 $ \\
\hline
&$10^{3}$  & $ 0.477$ & $2.094$ & $73.530$ & $0.529 $ \\
\hline
&$10^{5}$  & $ 0.486$ & $2.088$ & $132.420$ & $0.522 $ \\
\hline
&$10^{7}$  & $ 0.477$ & $2.088$ & $191.320$ & $0.488 $ \\
\hline
&$10^{9}$  & $ 0.273$ & $2.086$ & $200.000$ & $0.287 $ \\
\hline
\hline
 \UCIc  &$\bla_{init}$ & $\Rdeux$ & Time & \iter & $\Rdeux_{val}$  \\
\hline
\hline
 &  0.0  & $ -0.091$ & $0.052$ & $0.010$ & $-0.088 $ \\
\hline
&$10^{-3}$  & $ 0.761$ & $5.042$ & $97.970$ & $0.775 $ \\
\hline
&$10^{-1}$  & $ 0.795$ & $5.009$ & $66.370$ & $0.807 $ \\
\hline
&$10$  & $ 0.844$ & $4.989$ & $161.160$ & $0.847 $ \\
\hline
&$10^{3}$  & $ 0.843$ & $4.974$ & $194.550$ & $0.844 $ \\
\hline
 &  $\boldsymbol{10^4}$ & $ 0.814$ & $19.208^{\textbf{*}}$ & $197.100$ & $0.814 $ \\
\hline
&$10^{5}$  & $ 0.774$ & $4.966$ & $197.510$ & $0.775 $ \\
\hline
&$10^{7}$  & $ 0.711$ & $4.956$ & $198.600$ & $0.710 $ \\
\hline
&$10^{9}$  & $ 0.614$ & $4.942$ & $198.830$ & $0.613 $ \\
\hline
\end{tabular}
\end{table}



\paragraph{Ablation study.}
We ran our ablation study (Table \ref{tab:ablation}) in the regression setting on the same 3 datasets (\UCIa, \UCIb, \UCIc). We repeated each experiment over 100 random $train$/$test$ splits. All the results presented here correspond to the architecture of \MLRb{} and \BMLRb{} with hyperparameters fixed as in Table~\ref{tab:architectures1}.

A standard \RNN2 (\FFNN{} with $2$ wide layers $J=2^{10}$) cannot be trained efficiently on  small datasets as the \FFNN{} instantly memorizes the entire dataset. This cannot be alleviated through bagging at all. Note also its lower overall performance on the complete benchmark.

Applying Ridge on the last hidden layer allows an extremely overparametrized \FFNN{} to learn but its generalization performance is still far behind the gold standard \RF{}. However, when using bagging with ten such models, we reach very competitive results, underlying the potential of the \MLR{} approach.

The random permutations component gives a larger improvement than Structured Dithering. However, when using both ingredients together, a single \NN{} can reach or even outperform the gold-standard methods on most datasets. Furthermore, the improvement yielded by using bagging ($0.062$) is still of the same order of magnitude as the one we got when we applied permutations on top of Ridge to the \FFNN{} ($0.043$). This means these two ingredients (permutations and Structure Dithering) are not just simple variance reduction techniques but actually generate more sophisticated models.

\begin{table}[H]
\caption{Ablation Study in Regression.}
\label{tab:ablation}
	\centering
	\footnotesize
\begin{tabular}{|l||c|c|}
\hline
Step  &  Mean $\Rdeux$  & Bagging $\Rdeux$ \\
\hline
\hline
\RNN2  & $ -0.081  \pm  0.173 $ & $ -0.046  \pm  0.169 $ \\
\hline
\FFNN{}+ Ridge  & $ 0.321  \pm  0.081 $ & $ 0.394  \pm  0.052 $ \\
\hline
\FFNN{}+ Ridge + Struct. Dithering  & $ 0.323  \pm  0.075 $ & $ 0.400  \pm  0.048 $ \\
\hline
\FFNN{}+ Ridge + Permut. & $ 0.364  \pm  0.050 $ & $ 0.432  \pm  0.035 $ \\
\hline
\MLR{}   & $ 0.371  \pm  0.024 $ & $ 0.433  \pm  0.000 $ \\
\hline
\end{tabular}
\end{table}
\subsection{Other hyperparameters.} 

The impact of the other hyperparameters on the \MLR{} method is discussed below.


\paragraph{\Dither.}
At each iteration, we draw and add i.i.d. gaussian noise $\cN(0,\tilde{\sigma}^2\I)$ on the target $\bY$ in the regression setting. In Table~\ref{tab:Dithering}, we see that adding a small amount of noise improves performances. We performed our main experiments with $\tilde{\sigma} = 0.03$ as this value works well with standard \FFNN{}. But here again, we may improve generalization performance by considering $\tilde{\sigma}$ as an hyperparameter to be tuned. Rather unsurprisingly, applying dithering has no impact on runtime per iteration or on the value of $\bla_{init}$.



\begin{table}[H]
\caption{Dithering dependence : label noise scale. $\boldsymbol{\ast}$: computation time was obtained with a \Portable.}
\label{tab:Dithering}
	\centering
\begin{tabular}{|c|c|c|c|c|c|c|}
\hline
\UCIa & $\tilde{\sigma}$  &  $\Rdeux$ & Time & \iter & $\Rdeux_{val}$ & $\bla_{init}$  \\
\hline
\hline
 &  0  & $ 0.347$ & $3.104$ & $116.490$ & $0.483$ & $220.900 $ \\
\hline
 &  0.01  & $ 0.351$ & $3.110$ & $114.560$ & $0.486$ & $220.900 $ \\
\hline
 &  $\boldsymbol{0.03}$  & $ 0.347$ & $5.560^{\textbf{*}}$ & $116.190$ & $0.484$ & $216.235 $ \\
\hline
 &  0.1  & $ 0.354$ & $3.111$ & $113.720$ & $0.490$ & $215.104 $ \\
\hline
 &  0.3  & $ 0.353$ & $3.108$ & $119.300$ & $0.502$ & $216.367 $ \\
\hline
\UCIb  & $\tilde{\sigma}$ & $\Rdeux$ & Time & \iter & $\Rdeux_{val}$ & $\bla_{init}$  \\
\hline
\hline
 &  0 & $ 0.475$ & $3.250$ & $76.830$ & $0.527$ & $774.264 $ \\
\hline
 &  0.01  & $ 0.474$ & $3.258$ & $68.680$ & $0.527$ & $774.264 $ \\
\hline
 &  $\boldsymbol{0.03}$  & $ 0.473$ & $9.709^{\textbf{*}}$ & $72.800$ & $0.527$ & $774.264 $ \\
\hline
 &  0.1  & $ 0.474$ & $3.258$ & $70.860$ & $0.528$ & $774.264 $ \\
\hline
 &  0.3  & $ 0.474$ & $3.258$ & $68.620$ & $0.532$ & $774.264 $ \\
\hline
\UCIc  & $\tilde{\sigma}$ & $\Rdeux$ & Time & \iter & $\Rdeux_{val}$ & $\bla_{init}$  \\
\hline
\hline
 &  0  & $ 0.813$ & $8.554$ & $197.430$ & $0.814$ & $10000.001 $ \\
\hline
 &  0.01  & $ 0.814$ & $8.561$ & $197.240$ & $0.814$ & $10000.001 $ \\
\hline
 &  $\boldsymbol{0.03}$  & $ 0.814$ & $29.283^{\textbf{*}}$ & $197.100$ & $0.814$ & $10000.001 $ \\
\hline
 &  0.1  & $ 0.813$ & $8.561$ & $196.720$ & $0.814$ & $10000.001 $ \\
\hline
 &  0.3  & $ 0.812$ & $8.567$ & $196.220$ & $0.813$ & $10000.001 $ \\
\hline
\end{tabular}
\end{table}

\paragraph{Width.}
Most notably, Table~\ref{tab:Width} reveals that wide architectures (large $J$) usually provide better generalization performance. We recall that for standard \RNN{} trained without \MLR, wider architectures are more prone to overfitting. Table \ref{tab:Width} also reveals that larger architectures work better for bigger datasets like \UCIc. For small datasets, $J=2^{10}$ provides good generalization performance for smaller runtime. When the width parameter exceeds GPU memory, parallelization is lost and we observe a dramatic increase in computational time.






%
%
\begin{table}[H]
\caption{Width dependence.}
\label{tab:Width}
	\centering
\begin{tabular}{|c|c|c|c|c|c|c|}
\hline
\UCIa  &$J$ & $\Rdeux$ & Time & \iter & $\Rdeux_{val}$ &
$\bla_{init}$  \\
\hline
\hline
&$2^{4}$  & $ 0.184$ & $0.705$ & $162.120$ & $0.284$ & $210.307 $ \\
\hline
&$2^{6}$ & $ 0.276$ & $0.751$ & $160.030$ & $0.364$ & $211.555 $ \\
\hline
&$2^{8}$  & $ 0.325$ & $0.905$ & $135.400$ & $0.431$ & $205.351 $ \\
\hline
&$\boldsymbol{2^{10}}$  & $ 0.344$ & $2.201$ & $113.610$ & $0.484$ & $222.455 $ \\
\hline
&$2^{12}$  & $ 0.322$ & $15.796$ & $94.180$ & $0.503$ & $220.900 $ \\
\hline
\hline
 \UCIb   &$J$ & $\Rdeux$ & Time & \iter & $\Rdeux_{val}$ &
$\bla_{init}$  \\
\hline
\hline
&$2^{4}$  & $ 0.367$ & $0.724$ & $184.540$ & $0.379$ & $678.097 $ \\
\hline
&$2^{6}$ & $ 0.442$ & $0.743$ & $157.840$ & $0.471$ & $628.464 $ \\
\hline
&$2^{8}$  & $ 0.467$ & $0.907$ & $115.510$ & $0.512$ & $774.264 $ \\
\hline
&$\boldsymbol{2^{10}}$  & $ 0.470$ & $2.188$ & $71.790$ & $0.527$ & $774.264 $ \\
\hline
&$2^{12}$  & $ 0.460$ & $16.987$ & $37.210$ & $0.524$ & $774.264 $ \\
\hline
\hline
 \UCIc  &$J$ & $\Rdeux$ & Time & \iter & $\Rdeux_{val}$ &
$\bla_{init}$  \\
\hline
\hline
&$2^{4}$  & $ 0.622$ & $1.008$ & $200.000$ & $0.620$ & $9350.431 $ \\
\hline
&$2^{6}$ & $ 0.714$ & $1.134$ & $200.000$ & $0.713$ & $9927.827 $ \\
\hline
&$2^{8}$  & $ 0.773$ & $1.955$ & $199.880$ & $0.773$ & $10000.001 $ \\
\hline
&$\boldsymbol{2^{10}}$  & $ 0.825$ & $7.062$ & $198.240$ & $0.825$ & $10000.001 $ \\
\hline
&$2^{12}$  & $ 0.856$ & $54.121$ & $193.270$ & $0.857$ & $10000.001 $ \\
\hline
\end{tabular}
\end{table}

\paragraph{Batch size.} We added the \UCId{} of size $(n,d)=(43824,33)$ in this experiment in order to measure the impact of batch-size on a larger dataset
but this dataset was not included in the benchmark. 
In view of Table~\ref{tab:Batchsize}, our recommendation is very simple: "\textbf{\textit{As big as possible !}}". For small datasets this means using the entire train-set at each iteration, while GPU memory constraints rule out going beyond $2^{14}$ for large datasets.




\begin{table}[H]
	\centering
\caption{Batch size dependence}
\label{tab:Batchsize}
	\resizebox{\columnwidth}{!}{
\begin{tabular}{|c|c|c|c|c|c|c|}
\hline
 \UCIa &$\bs$  & $\Rdeux$ & Time & \iter & $\Rdeux_{val}$ & $\bla_{init}$  \\
\hline
\hline
&1  & $ -0.122$ & $4.375$ & $32.596$ & $0.014$ & $38.452 $ \\
\hline
&$2^{4}$  & $ 0.334$ & $5.194$ & $129.673$ & $0.520$ & $82.567 $ \\
\hline
&$2^{5}$  & $ 0.349$ & $5.194$ & $107.269$ & $0.517$ & $110.214 $ \\
\hline
&$2^{6}$  & $ 0.393$ & $5.352$ & $115.115$ & $0.500$ & $246.869 $ \\
\hline
&$\boldsymbol{\min(n,2^{14})=103}$    & $ 0.401$ & $5.238$ & $114.385$ & $0.499$ & $237.899 $ \\
\hline
\hline
\UCIb &$\bs$   & $\Rdeux$ & Time & \iter & $\Rdeux_{val}$ & $\bla_{init}$  \\
\hline
\hline
&1  & $ -0.014$ & $4.658$ & $38.020$ & $0.003$ & $290.923 $ \\
\hline
&$2^{4}$  & $ 0.415$ & $5.046$ & $148.680$ & $0.490$ & $158.198 $ \\
\hline
&$2^{5}$  & $ 0.459$ & $5.180$ & $141.260$ & $0.527$ & $204.647 $ \\
\hline
&$2^{6}$  & $ 0.474$ & $5.216$ & $128.820$ & $0.545$ & $253.497 $ \\
\hline
&$2^{7}$  & $ 0.477$ & $5.277$ & $103.270$ & $0.540$ & $388.678 $ \\
\hline
&$2^{8}$  & $ 0.478$ & $5.254$ & $97.010$ & $0.535$ & $774.264 $ \\
\hline
&$\boldsymbol{\min(n,2^{14})=546}$  & $ 0.475$ & $5.301$ & $72.470$ & $0.528$ & $774.264 $ \\
\hline
\hline
\UCIc &$\bs$   & $\Rdeux$ & Time & \iter & $\Rdeux_{val}$ & $\bla_{init}$  \\
\hline

\hline
&1  & $ 0.013$ & $4.536$ & $15.790$ & $0.013$ & $89.257 $ \\
\hline
&$2^{4}$  & $ 0.640$ & $5.317$ & $168.790$ & $0.642$ & $107.543 $ \\
\hline
&$2^{5}$  & $ 0.673$ & $5.375$ & $176.200$ & $0.675$ & $171.905 $ \\
\hline
&$2^{6}$  & $ 0.703$ & $5.360$ & $186.940$ & $0.705$ & $212.625 $ \\
\hline
&$2^{7}$  & $ 0.729$ & $5.413$ & $188.540$ & $0.730$ & $537.572 $ \\
\hline
&$2^{8}$  & $ 0.750$ & $5.447$ & $189.770$ & $0.751$ & $774.264 $ \\
\hline
&$2^{9}$  & $ 0.763$ & $5.460$ & $191.490$ & $0.765$ & $2641.979 $ \\
\hline
&$2^{10}$  & $ 0.785$ & $5.781$ & $192.900$ & $0.786$ & $2782.560 $ \\
\hline
&$2^{11}$  & $ 0.790$ & $6.908$ & $195.150$ & $0.791$ & $10000.001 $ \\
\hline
&$2^{12}$  & $ 0.813$ & $9.842$ & $196.930$ & $0.814$ & $10000.001 $ \\
\hline
&$\boldsymbol{\min(n,2^{14})=8760}$   & $ 0.824$ & $13.547$ & $197.800$ & $0.825$ & $10000.001 $ \\
\hline
\hline
\UCId &$\bs$  & $\Rdeux$ & Time & \iter & $\Rdeux_{val}$ &
$\bla_{init}$  \\
\hline

\hline
&1  & $ -0.003$ & $4.826$ & $65.470$ & $-0.003$ & $185.268 $ \\
\hline
&$2^{4}$  & $ 0.019$ & $5.344$ & $121.380$ & $0.021$ & $628.786 $ \\
\hline
&$2^{5}$  & $ 0.050$ & $5.629$ & $157.860$ & $0.052$ & $563.540 $ \\
\hline
&$2^{6}$  & $ 0.109$ & $5.659$ & $176.680$ & $0.110$ & $770.180 $ \\
\hline
&$2^{7}$  & $ 0.148$ & $5.646$ & $171.810$ & $0.149$ & $705.241 $ \\
\hline
&$2^{8}$  & $ 0.188$ & $5.678$ & $179.850$ & $0.190$ & $988.136 $ \\
\hline
&$2^{9}$  & $ 0.209$ & $5.771$ & $185.860$ & $0.211$ & $911.264 $ \\
\hline
&$2^{10}$  & $ 0.231$ & $6.165$ & $190.310$ & $0.233$ & $1001.640 $ \\
\hline
&$2^{11}$  & $ 0.254$ & $7.251$ & $192.960$ & $0.255$ & $1245.079 $ \\
\hline
&$2^{12}$  & $ 0.278$ & $10.199$ & $194.410$ & $0.279$ & $1376.753 $ \\
\hline
&$2^{13}$  & $ 0.298$ & $20.653$ & $195.740$ & $0.298$ & $2782.560 $ \\
\hline
&$\boldsymbol{\min(n,2^{14})=43824}$  & $ 0.316$ & $56.976$ & $193.520$ & $0.317$ & $3793.002 $ \\
\hline
\end{tabular}
}
\end{table}



\paragraph{Depth.} 

As we can see in Table~\ref{tab:Depth}, the optimal choice of the depth parameter seems to be data-dependent and significantly impacts the $\Rdeux$-score. This motivated the introduction of the bagging \MLR{} models that we described in the main paper. 
We consider only architectures of depth $L\in\cG_L=\{1,\,2,\,3,\, 4\}$ which reached state of the art results nonetheless. Going deeper is outside of the scope we set for this study, since it would probably require more careful and manual tuning of the hyperparameters on each dataset.


%
\begin{table}[H]
\footnotesize{
\caption{Depth dependence. Mean and standard deviation of $\Rdeux$-score over $100$ seeds.}
\label{tab:Depth}
	\centering
	\resizebox{\columnwidth}{!}{
\begin{tabular}{|l|c|c|c|c|c|c|}
\hline
Dataset & $n$ & $d$ & $\MLRa$& $\MLRb$& $\MLRc$& $\MLRd$\\
\hline
{Concrete Slump Test -1} & 103 & 8 & $0.940 \pm 0.029$ & $\boldsymbol{0.954 \pm 0.018}$ & $\boldsymbol{0.954 \pm 0.025}$ & $0.935 \pm 0.032$ \\
\hline
{Concrete Slump Test -3} & 103 & 8 & $\boldsymbol{0.399 \pm 0.132}$ & $0.313 \pm 0.171$ & $0.274 \pm 0.210$ & $0.226 \pm 0.149$ \\
\hline
{Concrete Slump Test -2} & 103 & 8 & $0.455 \pm 0.133$ & $0.453 \pm 0.159$ & $\boldsymbol{0.505 \pm 0.171}$ & $0.425 \pm 0.245$ \\
\hline
{Servo} & 168 & 24 & $0.836 \pm 0.031$ & $0.839 \pm 0.046$ & $\boldsymbol{0.854 \pm 0.043}$ & $0.842 \pm 0.049$ \\
\hline
{Computer Hardware} & 210 & 7 & $0.981 \pm 0.008$ & $0.984 \pm 0.008$ & $\boldsymbol{0.985 \pm 0.008}$ & $\boldsymbol{0.985 \pm 0.007}$ \\
\hline
{Yacht Hydrodynamics} & 308 & 33 & $0.952 \pm 0.021$ & $0.962 \pm 0.020$ & $\boldsymbol{0.965 \pm 0.020}$ & $0.960 \pm 0.021$ \\
\hline
{QSAR aquatic toxicity} & 546 & 34 & $0.448 \pm 0.081$ & $0.459 \pm 0.071$ & $0.458 \pm 0.090$ & $\boldsymbol{0.470 \pm 0.087}$ \\
\hline
{QSAR Bioconcentration classes}  & 779 & 25 & $\boldsymbol{0.672 \pm 0.042}$ & $0.668 \pm 0.049$ & $0.666 \pm 0.051$ & $0.670 \pm 0.051$ \\
\hline
{QSAR fish toxicity} & 909 & 18 & $\boldsymbol{0.590 \pm 0.043}$ & $0.586 \pm 0.037$ & $0.582 \pm 0.043$ & $0.579 \pm 0.046$ \\
\hline
{insurance} & 1338 & 15 & $\boldsymbol{0.839 \pm 0.024}$ & $0.837 \pm 0.026$ & $0.833 \pm 0.033$ & $0.832 \pm 0.028$ \\
\hline
{Communities and Crime} & 1994 & 108 & $0.679 \pm 0.031$ & $0.677 \pm 0.029$ & $\boldsymbol{0.680 \pm 0.030}$ & $\boldsymbol{0.680 \pm 0.027}$ \\
\hline
{Abalone R} & 4178 & 11 & $\boldsymbol{0.566 \pm 0.023}$ & $0.543 \pm 0.078$ & $0.523 \pm 0.163$ & $0.538 \pm 0.078$ \\
\hline
{squark automotive CLV training} & 8099 & 77 & $\boldsymbol{0.891 \pm 0.006}$ & $0.890 \pm 0.006$ & $0.889 \pm 0.006$ & $0.883 \pm 0.007$ \\
\hline
{Seoul Bike Sharing Demand} & 8760 & 15 & $0.850 \pm 0.009$ & $0.878 \pm 0.008$ & $\boldsymbol{0.901 \pm 0.008}$ & $0.893 \pm 0.008$ \\
\hline
{Electrical Grid Stability Simu} & 10000 & 12 & $0.937 \pm 0.003$ & $0.958 \pm 0.002$ & $\boldsymbol{0.963 \pm 0.002}$ & $0.955 \pm 0.002$ \\
\hline
{blr real estate prices} & 13320 & 2 & $0.514 \pm 0.012$ & $\boldsymbol{0.522 \pm 0.012}$ & $\boldsymbol{0.522 \pm 0.013}$ & ${0.521 \pm 0.013}$ \\
\hline
\end{tabular}}
}
\end{table}

\paragraph{Learning rate.} We used ADAM with default parameters except for the learning rate. Indeed, since the width and batch size we picked were outside of the usual ranges, we had to adjust the learning rate accordingly (Table~\ref{tab:architectures1}). We did not attempt to use another optimizer as ADAM worked well.

\paragraph{Scalability.} 
The main limitation is the size of the GPU VRAM with a current maximum of $32$G on the best available configuration.  We conducted these experiments on devices with either $8$ or $11$G$-$VRAM.
Recall that the cost for the inversion of a 
 $J\times J$ matrix is linear on a GPU thanks to parallelization whereas it is quadratic on a CPU.
 The runtime per iteration is almost constant since it depends mostly on width, depth, batch-size and number of permutations which are either fix or bounded (for batch-size).

\bibliographystyle{plain}
\bibliography{biblio}